\pdfoutput=1
\documentclass[10pt,letterpaper]{fairmeta}
\usepackage{amsmath,amssymb}
\usepackage[numbers,sort&compress]{natbib}
\usepackage{array,float,algorithm,algpseudocode}
\usepackage{url}
\usepackage{xspace}
\usepackage{adjustbox}

\definecolor{cvprblue}{rgb}{0.21,0.49,0.74}
\newcommand{\tbf}[1]{{\rmfamily\bfseries #1}}
\newcommand{\maketitlesupplementary}{%
  \par\noindent
  {\LARGE\sffamily\bfseries\color{metafg} Supplementary Material}\par
  \vspace{4pt}%
  {\noindent\color{metabg!45}\rule{\textwidth}{1pt}}\par
  \vspace{10pt}%
}
\newcommand{\headingperiod}[1]{}
\makeatletter
\def\theHALG@line{\thealgorithm.\arabic{ALG@line}}
\makeatother
\newcommand{\eg}{\emph{e.g.}\xspace}

\title{WorldOlympiad 2.0: Evaluating Embodied World Models with a Self-Evolving
Agent Harness and Continuous Action Evaluation}
\author[1,*]{Chengzhi Wang}
\author[1,*]{Yuke Zhao}
\author[1]{Zanyang Huang}
\author[2,\dagger]{Zeyu Zhang}
\author[1]{Shiyuan Zhang}
\author[1]{Weijie Wang}
\author[3]{Akide Liu}
\author[4,5]{Jiasheng Tang}
\author[2]{Yinghao Yu}
\author[1,4,\ddagger]{Bohan Zhuang}

\affiliation[1]{Zhejiang University}
\affiliation[2]{TRE, Alibaba Group}
\affiliation[3]{Monash University}
\affiliation[4]{DAMO Academy, Alibaba Group}
\affiliation[5]{Hupan Lab}

\contribution[*]{Equal contribution}
\contribution[\dagger]{Project lead}
\contribution[\ddagger]{Corresponding author}
\abstract{
Evaluating embodied world models requires assessing physical plausibility, 3D geometry, and instruction-aligned behavior across diverse simulators, embodiments, and rollout modes. Existing benchmarks rely on fixed task sets or passive-video judges, which cannot generate new physics-verified tasks or reliably adapt them across incompatible model interfaces—where subtle setup or conversion errors can bias evaluation scores. Meanwhile, prior motion metrics discretize rollouts: action-chunk metrics require explicit action heads and matched control rates, whereas frame judges and 2D tracks cannot recover continuous action trajectories or separate path errors from timing errors. We present \textbf{WorldOlympiad 2.0}, pairing a self-evolving evaluation harness with Continuous Action Evaluation (CAE) on WorldOlympiad's three-track framework. Unlike policy agents that optimize task execution, our harness runs and safeguards the evaluation itself: it synthesizes physics-checked canonical references, adapts them to native rollout interfaces, scores the resulting rollouts, and evolves by turning diagnosed failures into regression-tested updates to its tools, skills, and persistent records under frozen scoring rules. CAE recovers continuous action trajectories directly from RGB videos without action heads or aligned frame rates to score speed-invariant path fidelity and time-aligned velocity-vector fidelity. We evaluate five models on 500 cases each from a pool of 1,100, for 2,500 rollouts in total. LingBot-VA leads the shared dual-arm set, whereas GR00T’s simulator-executed rollouts score highly overall yet lowest on CAE—showing that visual ratings can mask severe trajectory errors. Newly collected real-robot demonstrations and rollout videos are included in the supplementary material.
}

\date{\today}
\metadata[Website]{\url{https://alibaba-damo-academy.github.io/WorldOlympiad/}}
\metadata[Code]{\url{https://github.com/alibaba-damo-academy/WorldOlympiad}}
\metadata[Email]{\email{bohan.zhuang@gmail.com}}

\begin{document}
\maketitle
\thispagestyle{firstheader}
\begin{figure}[t]

\centering
\includegraphics[trim=3bp 0 3bp 0,clip,width=\textwidth]{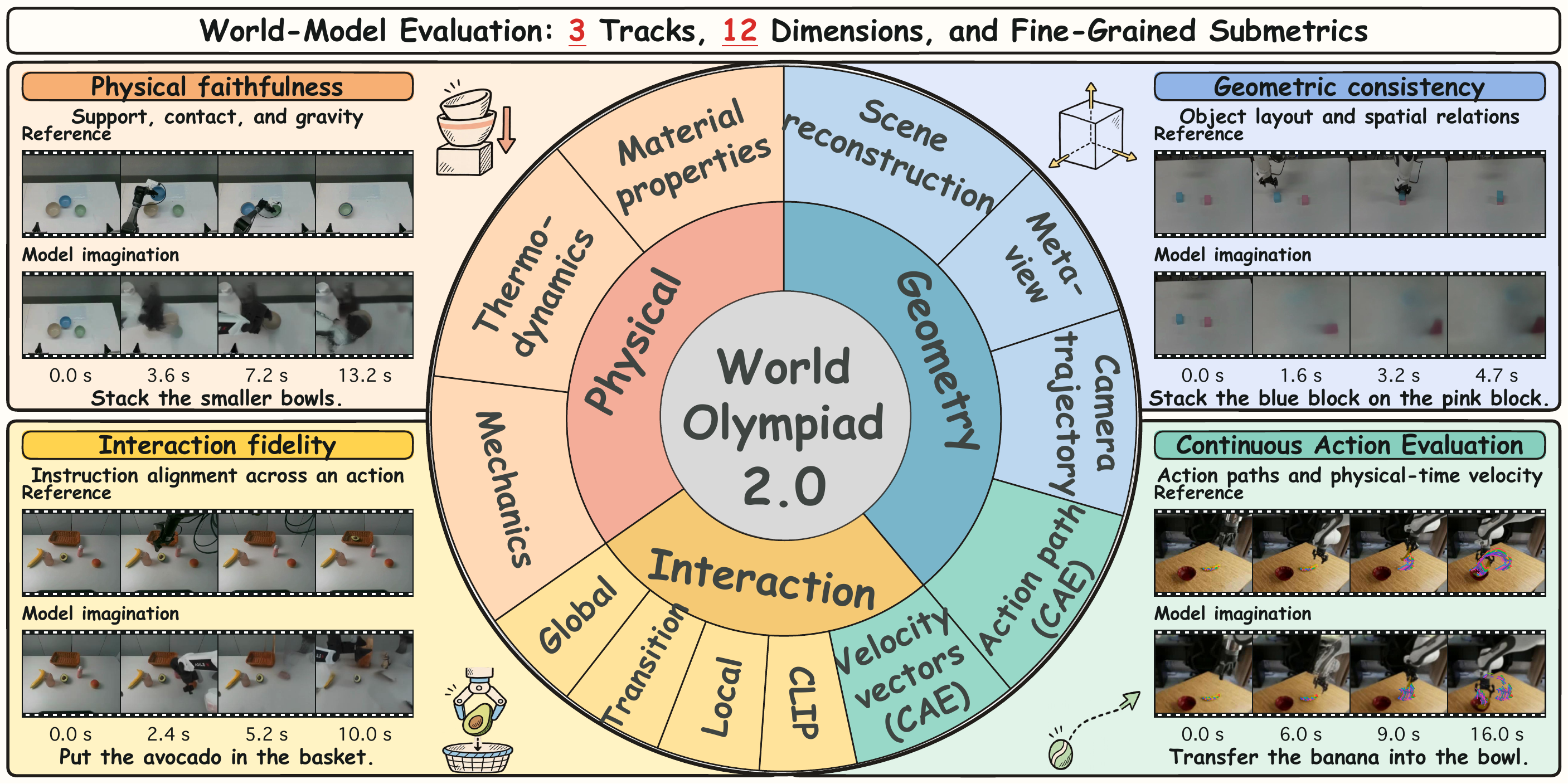}\par
\captionsetup{position=bottom,skip=8pt,hypcap=false}
\captionof{figure}{\textbf{Three-track world-model evaluation.} For continuous action trajectories, CAE adds path fidelity to geometry and velocity-vector fidelity to interaction. The 12 dimensions include thermodynamics from WorldOlympiad~\cite{zhao2026worldolympiad}. No episode triggered a thermodynamic rule.}
\captionsetup{hypcap=true}
\label{fig:teaser}\label{fig:evaluation_landscape}

\end{figure}

\section{Introduction}
\label{sec:intro}

Embodied world models support robot control by predicting how scenes evolve under actions~\cite{agarwal2026cosmos,chi2025wow,team2026openworldlib,wang2026worldr1,wang2026latentspatial}. Recent systems span action-free video generators paired with flow-based controllers~\cite{dharmarajan2026dream2flow,li2026novaflow}, video backbones fine-tuned as visuomotor policies~\cite{kim2026cosmos,ye2026world}, and joint video--action world models~\cite{li2026causal,yuan2026fast,huang2026long,zhou2026tau_0,geng2026trajwam}. Fairly evaluating all of them requires comparing the behaviors they predict or execute across incompatible simulators, embodiments, cameras, and rollout modes.

Existing efforts address parts of this problem across three categories: (i)~\emph{static world-model benchmarks} assess visual quality, physical commonsense, 2D motion tracks, or fixed simulator tasks~\cite{huang2024vbench,liu2024evalcrafter,bansal2025videophy,zhao2026worldolympiad,yue2025ewmbench,shang2026worldarena,jiang2026robowm}; (ii)~\emph{procedural task generators} synthesize simulation-ready scenes and demonstrations to expand task diversity beyond fixed datasets~\cite{mandlekar2023mimicgen,wang2023robogen,chen2025robotwin,wang2026scenecode,pfaff2026scenesmithagenticgenerationsimulationready}; and (iii)~\emph{automated evaluation and harness frameworks} unify static policy wrappers~\cite{choi2026vla}, run tool-calling agents to judge passive videos~\cite{tian2026open,chen2026harnesseval}, or evolve agent harnesses from execution feedback~\cite{lee2026meta,wang2026harness,wang2026harnesspai}.

Yet evaluating heterogeneous embodied world models in manipulation exposes two unresolved bottlenecks. First, scaling across simulators, robot embodiments, and rollout protocols creates a trade-off between \emph{task coverage} and \emph{benchmark validity} that existing harnesses do not solve: static wrappers~\cite{choi2026vla} cannot generate new physical references; passive-video evaluators~\cite{tian2026open,chen2026harnesseval} lack physical execution and pipeline repair, allowing invalid collision meshes or inaccurate goal checks to bias evaluation scores; and existing harness optimizers~\cite{lee2026meta,wang2026harness,wang2026harnesspai} modify the \emph{evaluated policy} to boost task performance rather than repairing \emph{benchmark infrastructure} under frozen scoring rules (Section~\ref{sec:related}). Second, current motion metrics rely on discrete representations: discrete action-chunk metrics require explicit action heads and matched control rates, whereas sparsely sampled frame judges and 2D tracks cannot recover continuous action trajectories or separate path errors from timing errors, allowing discontinuous or drifting rollouts to receive high visual scores.

\begin{figure*}[t]
\centering\includegraphics[width=\textwidth]{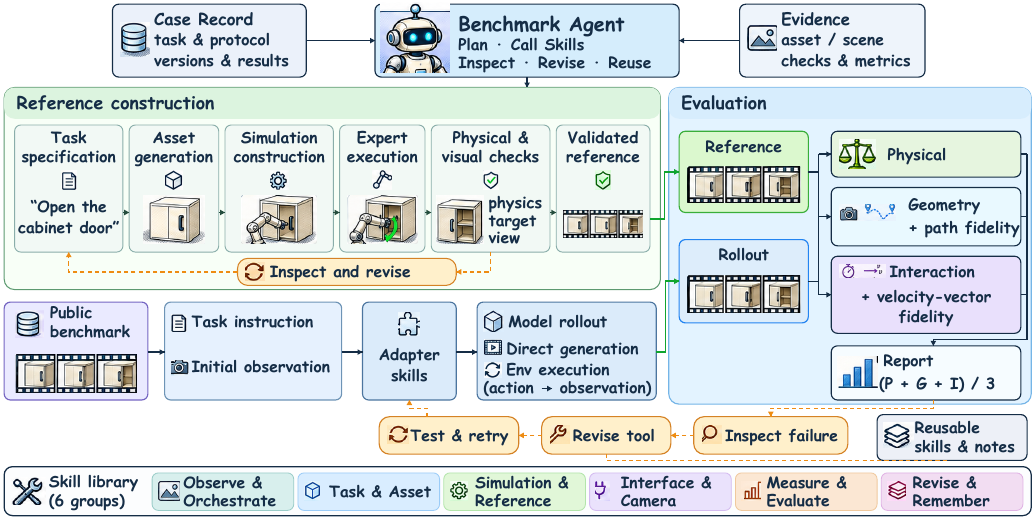}
\caption{\textbf{Agentic evaluation workflow.} An agent loop invokes reusable skills to construct canonical references, adapt them to native model protocols, and score rollouts. When infrastructure or validation checks fail, the agent diagnoses the trace, updates tools, skill instructions, or cross-round records, and verifies repairs via regression tests before reuse.}
\label{fig:method_overview}
\end{figure*}

To overcome both bottlenecks, we introduce \textbf{WorldOlympiad 2.0}, pairing a self-evolving evaluation harness with Continuous Action Evaluation (CAE) on WorldOlympiad's three-track framework~\cite{zhao2026worldolympiad} (Fig.~\ref{fig:teaser}). 
%Instead of maintaining error-prone per-model scripts, we run an \emph{agent loop} that invokes reusable skills to synthesize and physics-check simulation tasks alongside public episodes into standardized \emph{canonical references}, and then adapts each reference into every model's native input protocol (Fig.~\ref{fig:method_overview}). 
Our \emph{agent loop} uses reusable skills to construct and validate simulated references, incorporate public episodes, and score rollouts. It adapts each case to the native interfaces of models with compatible robot embodiments (Fig.~\ref{fig:method_overview}).
The harness evolves as it evaluates: when an asset, adapter, or goal predicate fails, the agent diagnoses the trace and updates its tool scripts, skill workflows, and persistent records of simulator conventions, interface specifications, and verified repairs, promoting only regression-tested updates while task goals and scoring rules stay locked. Complementing this harness, CAE recovers continuous action trajectories from RGB videos and measures speed-invariant path fidelity and time-aligned velocity-vector fidelity to augment the geometric and interaction tracks.

\begin{figure}[t]
\centering\includegraphics[width=\linewidth]{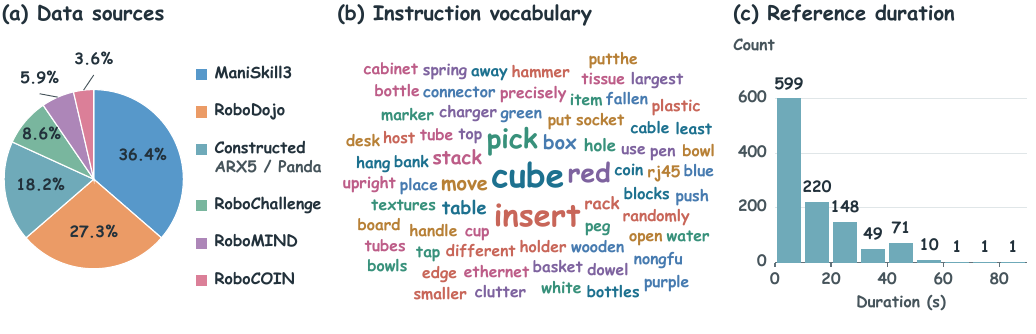}
\caption{\textbf{Reference-pool statistics.} Source distribution, instruction vocabulary, and native durations across 1,100 cases (200 constructed).}
\label{fig:data_sources}
\end{figure}

Our contributions are threefold:\nopagebreak[4]
\begin{itemize}
\setlength{\itemsep}{3pt}
\item \textbf{Self-evolving evaluation harness.} We build an evaluation agent that runs the full benchmark loop---constructing physics-verified references, adapting them to native model interfaces, executing rollouts, and scoring them---and evolves as it evaluates: each diagnosed pipeline failure becomes a regression-tested update to its tools, skills, and persistent records. Verified fixes thus accumulate across rounds and extend coverage to new simulators and embodiments, while task goals and scoring rules remain fixed.
\item \textbf{Continuous Action Evaluation (CAE).} We propose a model-agnostic scheme that recovers continuous action trajectories from RGB videos alone and scores progress-aligned path fidelity and time-aligned velocity-vector fidelity, placing video-only and action-producing models on a common motion scale at any frame or control rate while separating spatial errors from timing errors.
\item \textbf{Unified benchmark and empirical insights.} From the 1,100-case reference pool in Fig.~\ref{fig:data_sources}, each of five models runs on 500 compatible cases (400 public and 100 constructed), giving 2,500 rollouts in total. GR00T has the highest overall score of 0.643 but the lowest CAE path and velocity scores (Table~\ref{tab:main_results}). Human and automatic rankings broadly agree across the five models (Spearman $\rho=0.90$), although human raters place LingBot-VA first and GR00T second (Table~\ref{tab:human_eval}).
\end{itemize}
\section{Related Work}
\label{sec:related}

\paragraph{World-model evaluation.}
As video world models increasingly incorporate explicit 3D geometric constraints and persistent spatial memory~\cite{wang2026worldr1,wang2026latentspatial}, general video-generation benchmarks (e.g., VBench~\cite{huang2024vbench,huang2025vbench++,zheng2025vbench}, EvalCrafter~\cite{liu2024evalcrafter}, Beyond-FVD~\cite{luo2025beyond}) assess visual quality, semantic alignment, and temporal consistency, whereas physics-focused video benchmarks (e.g., VideoPhy~\cite{bansal2025videophy,bansal2026videophy}, PhyWorldBench~\cite{gu2026phyworldbench}, T2VPhysBench~\cite{guo2025t2vphysbench}, PVBench~\cite{chen2026pvbench}, and others~\cite{meng2024towards,sanli2026physvidbench}) test physical commonsense and rule compliance. Closer to embodied settings, WorldModelBench and WorldScore~\cite{li2026worldmodelbench,duan2025worldscore} evaluate instruction following and controllability, WBench~\cite{ying2026wbench} targets multi-turn interaction, RoboTrustBench~\cite{li2026robotrustbench} examines manipulation trustworthiness, and WorldOlympiad~\cite{zhao2026worldolympiad} unifies physical faithfulness, geometric consistency, and interaction fidelity. In contrast to these fixed, discrete evaluation suites, WorldOlympiad 2.0 pairs a self-evolving harness that builds, adapts, and repairs cross-protocol references with continuous action-trajectory scoring.

\paragraph{Trajectory-based evaluation.}
Motion-aware world-model benchmarks still rely on 2D or discrete action proxies: EWMBench~\cite{yue2025ewmbench} compares framewise 2D end-effector tracks via Hausdorff distance and dynamic time warping, WorldArena and WorldArena 2.0~\cite{shang2026worldarena,shang2026worldarena2} combine 2D trajectory checks with downstream policy metrics, and RoboWM-Bench~\cite{jiang2026robowm} maps generated rollouts to discrete actions to test simulator executability. Meanwhile, point trackers recover dense temporal correspondences~\cite{karaev2025cotracker3,koppula2024tapvid}, and video-to-action methods such as Dream2Flow~\cite{dharmarajan2026dream2flow} use predicted motion to plan robot trajectories~\cite{chi2025wow,li2026novaflow}. In contrast, our CAE recovers continuous action trajectories from RGB videos for \emph{model-agnostic evaluation}, decoupling spatial path geometry from physical-time velocity without requiring action heads, simulator states, or matched frame rates.

\paragraph{Agentic evaluation and harness evolution.}
Procedural generators and feed-forward 3D reconstructors synthesize simulation-ready scenes and demonstrations~\cite{mandlekar2023mimicgen,nasiriany2024robocasa,chen2025robotwin,wang2023robogen,wang2026scenecode,pfaff2026scenesmithagenticgenerationsimulationready,wang2026trisplat,wang2026feedforward}, while LLM agents combine tool use, self-reflection, and skill libraries~\cite{yao2022react,shinn2023reflexion,NEURIPS2023_91edff07,wang2023voyageropenendedembodiedagent} and increasingly coordinate world-state-centric video generation~\cite{wang2026anytime}. Recent evaluation frameworks either wrap static policy--benchmark APIs (vla-eval~\cite{choi2026vla}) or use agents to score passive videos without physical simulation (UniVBench~\cite{wei2026univbench}, Open Evaluation Agent~\cite{tian2026open}, HarnessEval-W~\cite{chen2026harnesseval}). Separately, harness-evolution methods (Meta-Harness~\cite{lee2026meta}, RiseHarness~\cite{wang2026harness}, HarnessPAI~\cite{wang2026harnesspai}, and related systems~\cite{zhang2026hyperagents,xu2026harness,yao2026harness}) optimize the \emph{evaluated agent's} prompts, memory, or control code to improve task scores—altering the tested system itself. In contrast, our harness evolves the \emph{benchmark infrastructure} rather than the evaluated models, repairing faulty simulation assets, native adapters, and goal predicates via regression tests under frozen tasks and scoring rules.
\section{WorldOlympiad 2.0}
\label{sec:worldolympiad2}

WorldOlympiad 2.0 pairs a self-evolving evaluation harness (Section~\ref{sec:agentic_harness}) with Continuous Action Evaluation (CAE; Section~\ref{sec:cae}) within WorldOlympiad's three-track framework (Section~\ref{sec:evaluation_framework}): \emph{Physical Faithfulness} ($P$), \emph{Geometric Consistency} ($G$), and \emph{Interaction Fidelity} ($I$). Each case specifies a language instruction, initial observations, a target object, and a reference video $V$ over a fixed media-time interval $[0,T]$; given only the instruction and initial observations, a world model produces rollout video $\widehat{V}$ on $[0,T]$ (hats mark rollout counterparts throughout, \eg, $\widehat{\mathbf{x}}, \widehat{\mathbf{v}}$).

\subsection{Agentic Evaluation and Harness Evolution}
\label{sec:agentic_harness}

\paragraph{Workflow overview.}
Comparing heterogeneous world models on shared tasks requires decoupling \emph{canonical reference creation} from \emph{model-specific execution}. Instead of maintaining separate benchmarks per interface, we run a closed-loop agent backed by reusable skills (\eg, procedural workflows), executable tools (\eg, simulation, rendering, and metric scripts), and persistent records (\eg, cross-round logs of simulator conventions, interface specifications, and verified repairs) across four checkpointed stages (Fig.~\ref{fig:method_overview}): scene construction, reference validation, native adaptation, and closed-loop repair. Checkpoints cache valid artifacts and halt scoring on unresolved pipeline errors.

\begin{figure*}[t]
\centering\includegraphics[width=\textwidth,trim=0bp .5bp 0bp 3.5bp,clip]{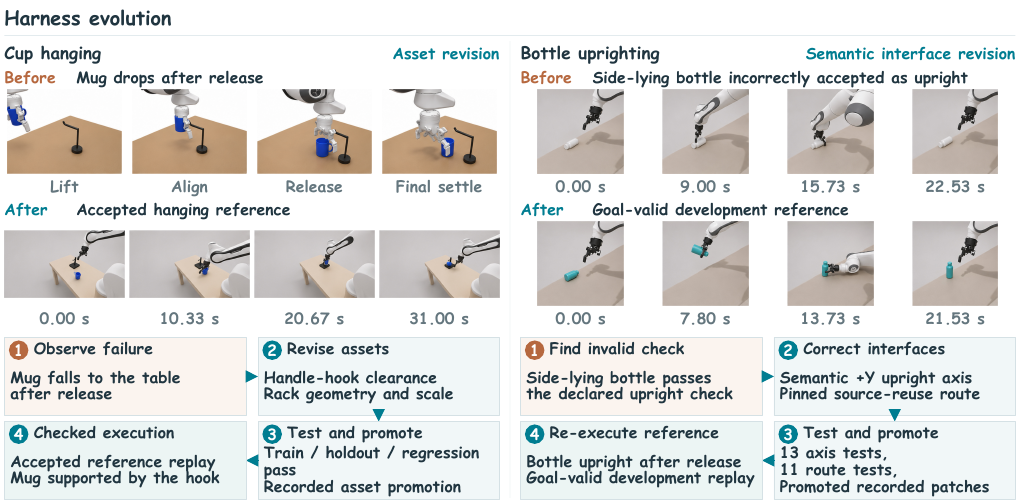}
\caption{\textbf{Harness evolution.} Verified closed-loop repairs for mug-hanging asset geometry (left) and bottle-uprighting goal checks (right).}
\label{fig:revisions}
\label{fig:harness_revision}
\end{figure*}

\paragraph{Scene construction and physical execution.}
For constructed cases, the agent converts a task specification into object assets, spatial layouts, initial states, and geometric goal predicates. Using SceneCode~\cite{wang2026scenecode} and parametric primitives, it builds the scene in PyBullet~\cite{coumans2016pybullet} with collision meshes and articulated joints, where expert controllers execute grasping, transport, reorientation, articulation, pushing, or tool use to record joint and 6-DoF object trajectories.

\paragraph{Reference validation and rendering.}
Because expert controllers can still slip or place objects unstably, automated goal predicates check joint progress, target pose, contact, and post-release static support, discarding failed candidates or routing them for repair. Passing trajectories are replayed in Blender or SAPIEN across standardized camera views to pair multi-view RGB videos $V$ with ground-truth states into a versioned \emph{canonical reference} (Appendix Algorithm~\ref{alg:synthesis}); curated public episodes enter directly at this stage.

\paragraph{Reference-to-model adaptation.}
From each canonical reference, model-specific adapters build native rollout packages by translating scene formats (\eg, SDF/URDF to USD), aligning coordinate frames and joint conventions, and assembling the required camera views, robot joint states, and history windows. Pre-inference checks verify loaded poses, joint limits, and target visibility before rollout execution (Appendix~\ref{sec:supp_native_conversion}).

\paragraph{Rollout execution, scoring, and harness evolution.}
In each round, the agent schedules pending case--model pairs from persistent records, runs native model inference to obtain $\widehat{V}$, and computes CAE and three-track scores (Appendix Algorithm~\ref{alg:evaluation}). We score rollouts even when the robot fails the task. Conversely, when asset, adapter, or goal-predicate errors occur, the agent inspects failure traces, updates the faulty tool script, skill workflow, or persistent record, and runs regression tests before re-executing the stage (Appendix~\ref{sec:supp_harness}; Fig.~\ref{fig:revisions}). Accepted updates persist across rounds, so later cases and newly added models reuse verified fixes rather than rediscovering them. All repairs keep task goals, model weights, scoring criteria, and intervals $[0,T]$ frozen, revalidating any updated reference before reuse.

\subsection{Continuous Action Evaluation}
\label{sec:cae}
\label{sec:trajectory_recovery}

\begin{figure*}[t]
\centering
\includegraphics[width=\textwidth]{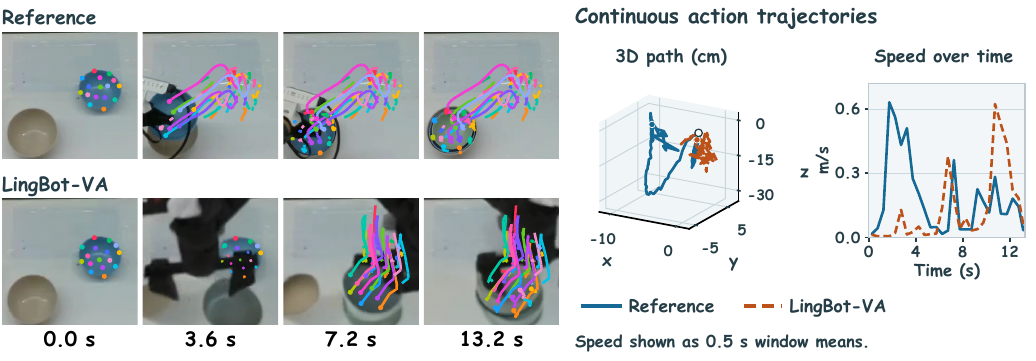}
\caption{\textbf{Continuous action trajectories in bowl stacking.} Left: tracks of 16 points on the blue bowl in the RoboChallenge~\cite{yakefu2025robochallenge} reference and LingBot-VA rollout. Right: their start-relative 3D paths and mean speeds in 0.5\,s windows. CAE scores native-time velocity vectors ($S_{\mathrm{path}}=0.603$, $S_{\mathrm{vel}}=0.297$).}
\label{fig:ee_continuous_motion}
\end{figure*}

From $V$ and $\widehat{V}$, CAE independently recovers continuous action trajectories $\mathbf{x}(t)$ and $\widehat{\mathbf{x}}(t)$ for $t \in [0,T]$ and decomposes motion agreement into \emph{path fidelity} $S_{\mathrm{path}} \in [0,1]$ (spatial alignment at matched progress, invariant to speed) and \emph{velocity-vector fidelity} $S_{\mathrm{vel}} \in [0,1]$ (direction and speed alignment at matched physical time).

\paragraph{Normalized signal similarity.}
\label{par:motion_notation}
For 3D signals $f(u)$ and $g(u)$ on interval $D$ with norm $\|f\|_D = (\int_D \|f(u)\|^2\,du)^{1/2}$ ($\|\cdot\|$ is the 3D Euclidean norm), we define
\begin{equation}
\mathcal{S}_D(f,g) = 1 - \frac{\|f - g\|_D}{\|f\|_D + \|g\|_D} \in [0,1],
\label{eq:similarity}
\end{equation}
with $\mathcal{S}_D(f,g) = 1$ if $\|f\|_D = \|g\|_D = 0$ (scoring $1$ when $f = g$ and $0$ when only one signal is zero everywhere).

\begin{figure*}[t]
\centering\includegraphics[width=\linewidth]{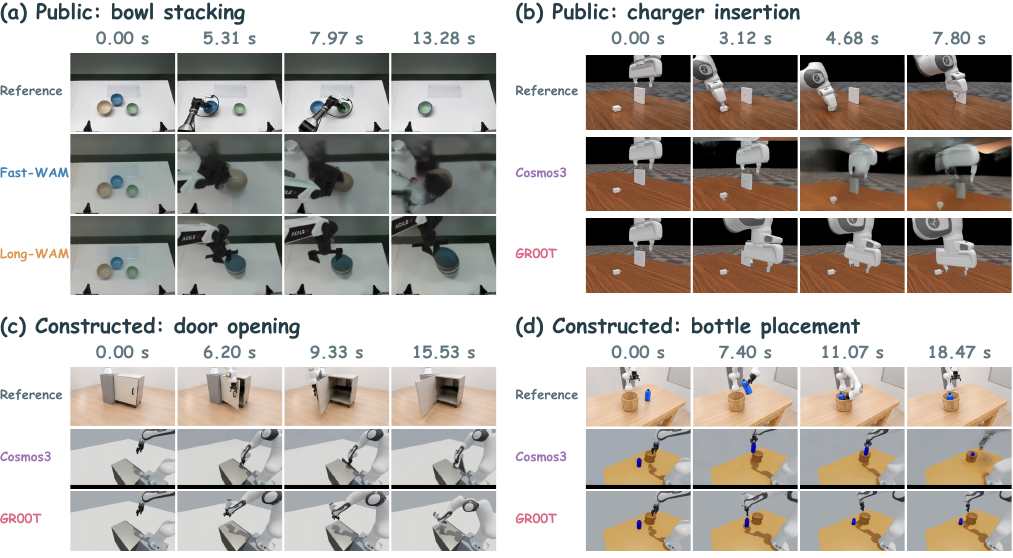}
\caption{\textbf{Model rollouts on public and constructed tasks.} Four matched timestamps for bowl stacking, charger insertion, door opening, and bottle placement (bowl-stacking trajectories in Fig.~\ref{fig:ee_continuous_motion}; more cases in Appendix~\ref{sec:supp_demos}).}
\label{fig:continuous_action_evaluation}
\label{fig:additional_rollouts}
\label{fig:main_model_comparisons}
\label{fig:main_constructed_comparisons}
\end{figure*}

\paragraph{Trajectory recovery and continuous representation.}
Following Dream2Flow~\cite{dharmarajan2026dream2flow}, we recover 3D object flow separately from each video by tracking points on the manipulated object and lifting them with monocular depth. Reference frame-zero intrinsics and depth set the metric scale. At each native timestamp $t_i$, we fit a rigid transform to points shared with the anchor frame and apply it to the anchor centroid. This gives a position $\mathbf{x}_i$ even when some points are occluded. Linear interpolation yields the target's \emph{continuous action trajectory} (Fig.~\ref{fig:ee_continuous_motion}). We use the same procedure for the reference and rollout, without post-hoc alignment (Appendix~\ref{sec:supp_trajectory_extraction}).

\paragraph{Path fidelity.}
For a trajectory of total length $L>0$, $s(t)$ is the fraction of its length traveled by time $t$, where $a(t)$ is the distance traveled so far:
\begin{equation}
a(t)=\int_0^t\|\dot{\mathbf{x}}(u)\|\,du,
\quad L=a(T),\quad s(t)=\frac{a(t)}{L}.
\label{eq:main_progress}
\end{equation}
We compare positions at the same $s\in[0,1]$. For example, $s=0.5$ is halfway along each trajectory by arc length. The first time it reaches that progress is $t(s)=\min\{\tau\in[0,T]:a(\tau)\ge sL\}$. All times within a pause give the same position. For a stationary trajectory ($L=0$), this definition gives $t(s)=0$ for all $s$. We define $\widehat{t}(s)$ analogously for the rollout and subtract the reference start from both paths:
\begin{equation}
\mathbf{q}(s)=\mathbf{x}(t(s))-\mathbf{x}(0),\;\; \widehat{\mathbf{q}}(s)=\widehat{\mathbf{x}}(\widehat{t}(s))-\mathbf{x}(0).
\label{eq:main_anchored_paths}
\end{equation}
Thus a static reference has $\mathbf{q}(s)=\mathbf{0}$, and a static rollout retains its initial offset $\widehat{\mathbf{x}}(0)-\mathbf{x}(0)$. Applying Eq.~\eqref{eq:similarity} on $D=[0,1]$ gives
\begin{equation}
S_{\mathrm{path}} = \mathcal{S}_{[0,1]}(\mathbf{q},\widehat{\mathbf{q}}) = 1 - \frac{\|\mathbf{q} - \widehat{\mathbf{q}}\|_{[0,1]}}{\|\mathbf{q}\|_{[0,1]} + \|\widehat{\mathbf{q}}\|_{[0,1]}}.
\label{eq:path}
\end{equation}

\paragraph{Velocity-vector fidelity.}
To evaluate temporal pacing, speed, and direction at matched physical time $t \in [0,T]$, we compute piecewise-constant velocities $\mathbf{v}(t) = (\mathbf{x}_{i+1}-\mathbf{x}_i)/(t_{i+1}-t_i)$ on each native frame interval $(t_i, t_{i+1})$ (and $\widehat{\mathbf{v}}(t)$ analogously), and apply Eq.~\eqref{eq:similarity} over $D=[0,T]$:
\begin{equation}
S_{\mathrm{vel}} = \mathcal{S}_{[0,T]}(\mathbf{v},\widehat{\mathbf{v}}) = 1 - \frac{\|\mathbf{v} - \widehat{\mathbf{v}}\|_{[0,T]}}{\|\mathbf{v}\|_{[0,T]} + \|\widehat{\mathbf{v}}\|_{[0,T]}}.
\label{eq:velocity}
\end{equation}
Traversing the same path in the same order preserves $S_{\mathrm{path}}=1$. Velocity fidelity decreases if the velocity vectors differ over a nonzero time interval. Both scores equal one exactly when the trajectories coincide throughout $[0,T]$. We integrate Eqs.~\eqref{eq:path} and~\eqref{eq:velocity} on the merged progress and time grids, respectively (Appendices~\ref{sec:geometric_consistency}--\ref{sec:supp_cae_properties}).

\subsection{Integrating CAE into the Three Tracks}
\label{sec:evaluation_framework}

We retain the physical score $P$ from WorldOlympiad~\cite{zhao2026worldolympiad}. Its geometric score $G_{\mathrm{base}}$ averages scene reconstruction, novel-view consistency, and camera-trajectory agreement. CAE adds path fidelity as a fourth equally weighted criterion:
\begin{equation}
G = 0.75\,G_{\mathrm{base}} + 0.25\,S_{\mathrm{path}}.
\label{eq:geometry}
\end{equation}
$I_{\mathrm{base}}$ combines local, transition, and global ratings with calibrated CLIP similarity. For episode $i$ with multiple chunks,
\begin{equation}
I_i = 0.80\,I_{\mathrm{base},i} + 0.20\,S_{\mathrm{vel},i}.
\label{eq:interaction}
\end{equation}
For single-chunk episodes, omitting the transition term increases the velocity weight to $0.20/0.76$ (Appendix~\ref{sec:supp_track_integration}). We report both CAE scores, all three tracks, and $S_{\mathrm{all}}=(P+G+I)/3$.
\section{Experiments}
\label{sec:experiments}

\subsection{Experimental Setup}
\label{sec:experimental_setup}

\begin{table}[t]
\centering\small
\renewcommand{\arraystretch}{1.14}
\setlength{\tabcolsep}{1.5pt}
\caption{\textbf{Main benchmark results and track components} (400 public and 100 constructed cases per model). Bold and underline indicate first and second place on shared dual-arm cases. All is the per-case mean of $P$, $G$, and $I$. Path and Vel.\ are CAE scores. Local denotes local interaction ($I_{\mathrm{local}}$), and Sem.\ denotes semantic alignment measured by calibrated CLIP similarity ($I_{\mathrm{clip}}$).}
\label{tab:main_results}
\begin{tabular*}{\textwidth}{@{\extracolsep{\fill}}lrrr rrrrr rrrrrr r@{}}
\toprule
\multirow{2}{*}{\textbf{Model}} & \multicolumn{3}{c}{\textbf{Physical faithfulness}} & \multicolumn{5}{c}{\textbf{Geometric consistency}} & \multicolumn{6}{c}{\textbf{Interaction fidelity}} & \multirow{2}{*}{\textbf{All}}\\
\cmidrule(lr){2-4}\cmidrule(lr){5-9}\cmidrule(lr){10-15}
 & Mech. & Mat. & $P$ & Recon. & Meta & Cam. & Path & $G$ & Local & Trans. & Global & Sem. & Vel. & $I$ & \\
\midrule
\multicolumn{16}{@{}l}{\emph{Dual-arm (shared episode set)}}\\
Fast-WAM & 0.708 & 0.662 & 0.702 & 0.315 & 0.320 & 0.484 & 0.242 & 0.340 & 0.365 & 0.244 & 0.431 & 0.359 & 0.156 & 0.320 & 0.454\\
Long-WAM & 0.748 & 0.691 & \underline{0.726} & 0.365 & 0.363 & 0.588 & 0.243 & \tbf{0.390} & 0.403 & 0.369 & 0.510 & 0.361 & 0.157 & \underline{0.373} & \underline{0.496}\\
LingBot-VA & 0.744 & 0.732 & \tbf{0.735} & 0.374 & 0.377 & 0.556 & 0.246 & \underline{0.388} & 0.463 & 0.549 & 0.546 & 0.388 & 0.159 & \tbf{0.433} & \tbf{0.519}\\
\midrule
\multicolumn{16}{@{}l}{\emph{Single-arm (native protocols)}}\\
Cosmos3 & 0.774 & 0.577 & 0.685 & 0.372 & 0.350 & 0.265 & 0.382 & 0.342 & 0.394 & 0.235 & 0.231 & 0.421 & 0.183 & 0.289 & 0.439\\
GR00T & 0.915 & 0.932 & 0.920 & 0.537 & 0.510 & 0.903 & 0.239 & 0.547 & 0.531 & 0.208 & 0.736 & 0.523 & 0.146 & 0.460 & 0.643\\
\bottomrule
\end{tabular*}
\end{table}

\paragraph{Models.}
We evaluate five embodied world models across dual- and single-arm manipulation setups: Fast-WAM~\cite{yuan2026fast}, Long-WAM~\cite{huang2026long}, LingBot-VA~\cite{li2026causal}, Cosmos3 Nano-Policy-DROID~\cite{agarwal2026cosmos}, and GR00T N1.6-DROID~\cite{khazatsky2024droid}. Each model runs under its native conditioning and rollout interface without modifying architectures or control rates (Appendix~\ref{sec:supp_cohorts}, Tables~\ref{tab:cohort_classes}--\ref{tab:native_interfaces}); GR00T rollouts record simulated execution of its predicted action chunks.

\paragraph{Data.}
From the 1,100-case reference pool (Fig.~\ref{fig:data_sources}; Appendix Table~\ref{tab:data_sources}), each model runs on 400 public episodes (160 grasping/manipulation, 160 insertion/alignment, and 80 stacking) and 100 embodiment-matched constructed cases (20 each for cabinet articulation, cup uprighting, cup hanging, tool use, and tissue-box pushing). Public episodes come from RoboDojo~\cite{chen2026robodojo}, ManiSkill3~\cite{tao2024maniskill3}, RoboCOIN~\cite{wu2025robocoin}, RoboChallenge/Table30~\cite{yakefu2025robochallenge}, and RoboMIND~\cite{wu2024robomind}. The three dual-arm models share the same 300 RoboDojo and 100 real episodes, while Cosmos3 and GR00T share 300 ManiSkill plus 100 real (Cosmos3) or 100 simulated (GR00T) episodes; newly collected real-robot demonstrations appear in Appendix~\ref{sec:real_robot_demonstrations} and the supplementary video.

\paragraph{Scoring.}
Visual tracks use Qwen3-VL-8B-Instruct~\cite{bai2025qwen3}, and CAE scores continuous action trajectories recovered from RGB. Using each video's native timestamps, we compare paths at matched progress and velocity vectors at matched elapsed time, without resampling either video to a common frame rate. All 2,500 runs complete, with failed task rollouts retained and scored normally. Geometry, interaction, and CAE cover all 500 cases per model; 14 constructed runs have no triggered physical rule and are marked N/A on $P$ and All (Appendix~\ref{sec:supp_applicability}, Table~\ref{tab:applicable_counts}).

\subsection{Main Benchmark Results}
\label{sec:main_results}

\begin{figure}[t]
\centering
\includegraphics[trim=0 2bp 0 2bp,clip,width=\textwidth]{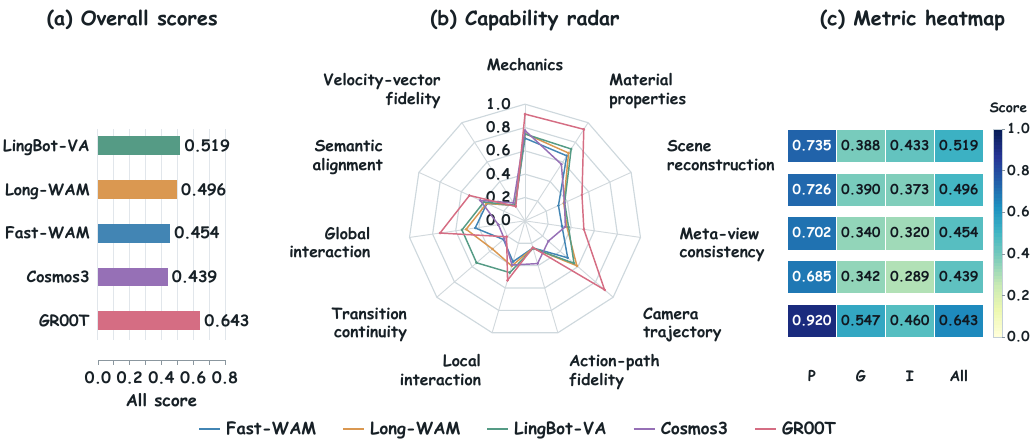}
\caption{\textbf{Evaluation profiles.} The radar shows 11 of the 12 dimensions in Fig.~\ref{fig:teaser}: no thermodynamic rule was triggered in 2,500 evaluations.}
\label{fig:evaluation_radar}
\end{figure}

\paragraph{Overall comparison.}
Table~\ref{tab:main_results} shows that LingBot-VA leads on the shared dual-arm set with an overall score of 0.519, followed by Long-WAM at 0.496 and Fast-WAM at 0.454. The largest gap is in interaction fidelity: 0.113 across models, compared with 0.050 for geometry and 0.033 for physical faithfulness. Transition scores follow the same ranking at 0.549, 0.369, and 0.244. Long-WAM leads narrowly on geometry, scoring 0.390 to LingBot-VA's 0.388, with the strongest camera-trajectory agreement at 0.588. On their respective single-arm protocols, GR00T scores 0.643 overall and Cosmos3 scores 0.439. Tables~\ref{tab:physical_rules_full}--\ref{tab:interaction_model_details} give the component scores.

\paragraph{Continuous action fidelity.}
Figure~\ref{fig:evaluation_radar} shows a different ranking for continuous action trajectories. Cosmos3 leads on both path and velocity fidelity at 0.382 and 0.183. GR00T scores 0.903 on camera-trajectory agreement and 0.736 on global interaction, but ranks last on path fidelity at 0.239, velocity fidelity at 0.146, and transition smoothness at 0.208. Its simulated rollouts preserve background appearance even when execution pauses between action chunks or the trajectory drifts. On the 500 shared dual-arm cases, mean scores differ by at most 0.004 on either CAE metric, despite larger differences in transition smoothness.

\paragraph{Public and constructed cases.}
Path fidelity drops on constructed tasks for four of the five models. Cosmos3 shows the largest drop, from 0.434 to 0.174. GR00T is the exception, rising slightly from 0.237 to 0.246. Velocity fidelity increases for all five under standardized simulation timing, from 0.127--0.173 on public episodes to 0.211--0.231 on constructed cases. LingBot-VA scores 0.580 overall on constructed tasks, followed by GR00T at 0.574 and Long-WAM at 0.542. Cup hanging is the hardest family for the dual-arm models, with scores of 0.372--0.537 under the post-release support criterion (Appendix Tables~\ref{tab:cae_source_split}--\ref{tab:synthetic_families}).

\subsection{Continuous Action Analysis}
\label{sec:cae_analysis}

Controlled perturbations confirm that $S_{\mathrm{path}}$ and $S_{\mathrm{vel}}$ isolate complementary errors (Appendix~\ref{sec:supp_controls}, Table~\ref{tab:controlled}): mid-rollout pauses preserve $S_{\mathrm{path}}$ while lowering $S_{\mathrm{vel}}$ (1.000 vs.\ 0.677), constant spatial offsets preserve $S_{\mathrm{vel}}$ while penalizing $S_{\mathrm{path}}$ (0.857 vs.\ 1.000), halving trajectory extent yields 0.667 on both, and reversed motion scores 0.000. Each perturbation keeps individual frames visually plausible, so frame-level judges cannot make these distinctions. In the bowl-stacking example in Fig.~\ref{fig:ee_continuous_motion}, CAE compares the reference and LingBot-VA trajectories at their native timestamps, yielding $S_{\mathrm{path}}=0.603$ and $S_{\mathrm{vel}}=0.297$.

\begin{table}[b]
\centering
\begin{minipage}[t]{.48\textwidth}
\centering\small
\renewcommand{\arraystretch}{1.12}
\captionsetup{position=top,justification=raggedright,singlelinecheck=false,skip=6pt}
\setlength{\tabcolsep}{2pt}
\caption{\textbf{Applicable physical-rule sub-scores.} Mechanics (Grav.--Impact) and Material (Hard.) means; \textemdash: N/A.}
\label{tab:physical_rules_full}
\label{tab:physical_subscores}
\label{tab:physical_rule_details}
\begin{tabular*}{\linewidth}{@{\extracolsep{\fill}}lrrrrr@{}}
\toprule
\textbf{Model} & \textbf{Grav.} & \textbf{Buoy.} & \textbf{Comp.} & \textbf{Impact} & \textbf{Hard.}\\
\midrule
Fast-WAM & 0.711 & 0.980 & 0.852 & 0.414 & 0.662\\
Long-WAM & 0.752 & 0.950 & 0.848 & 0.550 & 0.691\\
LingBot-VA & 0.748 & 0.980 & 0.875 & 0.505 & 0.732\\
\midrule
Cosmos3 & 0.779 & \textemdash & 0.550 & 0.543 & 0.577\\
GR00T & 0.928 & \textemdash & 0.912 & 0.704 & 0.932\\
\bottomrule
\end{tabular*}
\end{minipage}\hfill
\begin{minipage}[t]{.48\textwidth}
\centering\small
\renewcommand{\arraystretch}{1.12}
\setlength{\tabcolsep}{2pt}
\captionsetup{position=top,justification=raggedright,singlelinecheck=false,skip=6pt}
\caption{\textbf{Geometric sub-scores.} $G_{\mathrm{base}}$ is inherited geometry; Path is CAE fidelity and $G$ is the final track (Eq.~\eqref{eq:geometry}).}
\label{tab:geometry_model_details}
\label{tab:geometry_subscores}
\label{tab:inherited_geometry_detail}
\begin{tabular*}{\linewidth}{@{\extracolsep{\fill}}lrrrrrr@{}}
\toprule
\textbf{Model} & \textbf{Recon.} & \textbf{Meta} & \textbf{Cam.} & $G_{\mathrm{base}}$ & \textbf{Path} & $G$\\
\midrule
Fast-WAM & 0.315 & 0.320 & 0.484 & 0.373 & 0.242 & 0.340\\
Long-WAM & 0.365 & 0.363 & 0.588 & 0.439 & 0.243 & 0.390\\
LingBot-VA & 0.374 & 0.377 & 0.556 & 0.436 & 0.246 & 0.388\\
\midrule
Cosmos3 & 0.372 & 0.350 & 0.265 & 0.329 & 0.382 & 0.342\\
GR00T & 0.537 & 0.510 & 0.903 & 0.650 & 0.239 & 0.547\\
\bottomrule
\end{tabular*}
\end{minipage}
\end{table}

\subsection{Harness Evolution}
\label{sec:harness_results}

Closed-loop harness evolution repairs faulty assets and goal predicates prior to model scoring (Fig.~\ref{fig:revisions}). In cup hanging, after a released mug fell off the hook, the agent diagnosed insufficient handle-to-hook clearance, revised rack geometry and scale, and verified post-release stability across train, holdout, and regression checks. In bottle uprighting, where a side-lying bottle falsely passed an upright check, the agent fixed the $+Y$ upright-axis convention and asset-loading path, passing 24 regression checks (13 axis and 11 loading tests) before rebuilding a valid reference. Left unrepaired, the first fault would have discarded a feasible task and the second would have admitted a failed demonstration as ground truth, silently biasing every model's score on these cases. Across the benchmark, the harness resolved every recorded pipeline failure before scoring, requiring on average 24 repair attempts per constructed case but only 0.49 per public episode (Appendix~\ref{sec:supp_algorithms}).

\begin{table}[t]
\centering\small
\renewcommand{\arraystretch}{1.14}
\captionsetup{position=top,justification=raggedright,singlelinecheck=false,skip=6pt}
\setlength{\tabcolsep}{2.5pt}
\caption{\textbf{Interaction scoring components.} Global is the overall rollout rating. Long Range and Global Text are separate diagnostic ratings. Raw CLIP is uncalibrated cosine similarity. We renormalize weights for single-chunk episodes before computing the mean (Appendix~\ref{sec:supp_track_integration}).}
\label{tab:interaction_model_details}
\label{tab:interaction_subscores}
\label{tab:additional_interaction_diagnostics}
\begin{tabular*}{\textwidth}{@{\extracolsep{\fill}}llrrrrrrrrrr@{}}
\toprule
\textbf{Arm} & \textbf{Model} & \textbf{Local} & \textbf{Trans.} & \textbf{Global} & \textbf{Long Range} & \textbf{Global Text} & \textbf{Raw CLIP} & \textbf{Sem.} & $I_{\mathrm{base}}$ & \textbf{Vel.} & $I$\\
\midrule
\multirow{3}{*}{Dual-arm} & Fast-WAM & 0.365 & 0.244 & 0.431 & 0.446 & 0.416 & 0.272 & 0.359 & 0.367 & 0.156 & 0.320\\
 & Long-WAM & 0.403 & 0.369 & 0.510 & 0.531 & 0.488 & 0.272 & 0.361 & 0.435 & 0.157 & 0.373\\
 & LingBot-VA & 0.463 & 0.549 & 0.546 & 0.573 & 0.519 & 0.278 & 0.388 & 0.510 & 0.159 & 0.433\\
\midrule
\multirow{2}{*}{Single-arm} & Cosmos3 & 0.394 & 0.235 & 0.231 & 0.243 & 0.225 & 0.284 & 0.421 & 0.323 & 0.183 & 0.289\\
 & GR00T & 0.531 & 0.208 & 0.736 & 0.760 & 0.735 & 0.305 & 0.523 & 0.559 & 0.146 & 0.460\\
\bottomrule
\end{tabular*}
\end{table}

\subsection{Human Ratings}
\label{sec:human_eval}

Ten raters independently scored each applicable episode on the physical, geometric, and interaction tracks. We averaged their ratings for each episode and track, then used the same aggregation as for the automatic scores (Appendix~\ref{sec:supp_human}). Across the five model means, the human and automatic rankings agree on ranks three through five, with Spearman $\rho=0.90$. The top two are reversed: human raters place LingBot-VA first at 0.608 and GR00T second at 0.595 (Table~\ref{tab:human_eval}).

\begin{table}[H]
\centering\small
\begin{minipage}{0.88\textwidth}
\centering
\renewcommand{\arraystretch}{1.14}
\setlength{\tabcolsep}{6pt}
\captionsetup{position=top,skip=6pt}
\caption{\textbf{Human assessment.} Matched case- and track-level means (Appendix~\ref{sec:supp_human}); H (human rank) and A (automatic rank) denote model rankings across the five evaluated models, with $\Delta = \mathrm{H} - \mathrm{A}$.}
\label{tab:human_eval}
\begin{tabular*}{\linewidth}{@{\extracolsep{\fill}}lrrrrr@{}}
\toprule
\textbf{Model} & \textbf{Human mean} & \textbf{Auto.\ mean} & \textbf{Human rank (H)} & \textbf{Auto.\ rank (A)} & $\Delta$\\
\midrule
Fast-WAM & 0.487 & 0.454 & 4 & 4 & 0\\
Long-WAM & 0.512 & 0.496 & 3 & 3 & 0\\
LingBot-VA & 0.608 & 0.519 & 1 & 2 & $-1$\\
\midrule
Cosmos3 & 0.435 & 0.439 & 5 & 5 & 0\\
GR00T & 0.595 & 0.643 & 2 & 1 & 1\\
\bottomrule
\end{tabular*}
\end{minipage}
\end{table}

\section{Conclusion}
\label{sec:conclusion}

We presented \textbf{WorldOlympiad 2.0} to evaluate embodied world models beyond fixed simulators and discrete motion proxies~\cite{zhao2026worldolympiad}. Its \emph{self-evolving evaluation harness} turns benchmark construction, execution, and scoring into an agentic process that improves with use: each diagnosed failure becomes a regression-tested update to its tools, skills, and records, expanding coverage to new tasks, simulators, and embodiments under fixed scoring rules. Its \emph{Continuous Action Evaluation (CAE)} scores RGB-recovered continuous action trajectories on progress-aligned path and time-aligned velocity fidelity, placing video-only and action-producing models on a common motion scale. Consistent with human ratings ($\rho=0.9$), our results show that high visual scores often mask severe trajectory drift and execution pauses. We hope WorldOlympiad 2.0 serves as a living benchmark evolving alongside the world models it evaluates.

\paragraph{Limitations.}
Our tasks focus on tabletop manipulation. CAE depends on target visibility, segmentation, point tracking, and monocular depth. Camera motion can also affect the recovered trajectory because positions are expressed in the initial camera frame. CAE omits orientation and contact forces and may penalize valid alternative paths. Single-arm scores reflect each model's native rollout protocol and episode set (Appendix~\ref{sec:limitations}).

\FloatBarrier
{\small
\bibliographystyle{ieeenat_fullname}
\bibliography{main}
}
\clearpage
\appendix
\maketitlesupplementary
\setcounter{figure}{0}
\setcounter{table}{0}
\renewcommand{\thefigure}{\Alph{section}\arabic{figure}}
\renewcommand{\thetable}{\Alph{section}\arabic{table}}
\renewcommand{\theHfigure}{supp.\arabic{figure}}
\renewcommand{\theHtable}{supp.\arabic{table}}
% Unified supplementary source. All appendix prose, tables, and demonstrations
% are maintained in this file.
\newcommand{\suppTableHead}[1]{\textbf{#1}}
\raggedbottom

\section{Benchmark Data and Native Execution Protocols}
\label{sec:supp_sources}

\subsection{Reference Pool and Public Episode Splits}
\label{sec:supp_cohorts}

Tables~\ref{tab:data_sources} and~\ref{tab:cohort_classes} summarize the 1,100-case canonical reference pool and the evaluation split assigned to each model. The public pool contains 900 episodes (700 simulated and 200 real-world) converted directly into canonical references from curated demonstrations and native camera views. Each model is evaluated on 400 public episodes---balanced across 160 grasping/manipulation, 160 insertion/alignment, and 80 stacking tasks---together with 100 embodiment-matched constructed simulation cases.

\begin{table}[!ht]
\centering\small
\renewcommand{\arraystretch}{1.10}
\setlength{\tabcolsep}{6pt}
\caption{\textbf{Reference-pool composition and robot types.} Counts denote canonical references in the 1,100-case pool, grouped by source environment and robot arm configuration.}
\label{tab:data_sources}
\begin{tabular*}{\textwidth}{@{\extracolsep{\fill}}lllr@{}}
\toprule
\suppTableHead{Source} & \suppTableHead{Env.} & \suppTableHead{Robot / arm} & \suppTableHead{Count}\\
\midrule
RoboDojo & Sim. & ALOHA--AgileX, dual & 300\\
ManiSkill3 & Sim. & Panda, single & 400\\
RoboCOIN & Real & AgileX Cobot Magic, dual & 40\\
\multirow{2}{*}{RoboChallenge/Table30} & Real & ALOHA, dual & 60\\
 & Real & Franka, single & 35\\
RoboMIND & Real & Franka, single & 65\\
\midrule
\textbf{Public subtotal} & & & \tbf{900}\\
\midrule
\multirow{2}{*}{\textbf{Ours constructed}} & \multirow{2}{*}{Sim.} & ARX5, dual & 100\\
 & & Panda, single & 100\\
\midrule
\textbf{Constructed subtotal} & & & \tbf{200}\\
\textbf{Total reference pool} & & & \tbf{1,100}\\
\bottomrule
\end{tabular*}
\end{table}

\begin{table}[!ht]
\centering\small
\renewcommand{\arraystretch}{1.10}
\setlength{\tabcolsep}{6pt}
\caption{\textbf{Native embodiments and public-episode splits.} Groups A and B denote shared dual-arm and single-arm core sets, respectively.}
\label{tab:cohort_classes}
\begin{tabular*}{\textwidth}{@{\extracolsep{\fill}}lll@{}}
\toprule
\suppTableHead{Model} & \suppTableHead{Embodiment} & \suppTableHead{Public episodes}\\
\midrule
Fast-WAM & Dual-arm & A: 300 simulated + 100 real\\
Long-WAM & Dual-arm & A: 300 simulated + 100 real\\
LingBot-VA & Dual-arm & A: 300 simulated + 100 real\\
\midrule
Cosmos3 & Single-arm, Franka & B: 300 simulated + 100 additional real\\
GR00T & Single-arm, Panda & B: 300 simulated + 100 additional simulated\\
\bottomrule
\end{tabular*}

\end{table}

Specifically, the shared dual-arm simulation set (Group A, 300 episodes) covers pick-anything (50), general pickup (50), dual-bottle pickup (20), coin deposit (40), single-tube insertion (50), multi-tube insertion (30), block stacking (50), and toy stacking (10), paired with 100 real-world episodes (40 RoboCOIN fruit-basket, 40 Table30 cable-insertion, and 20 Table30 bowl-stacking; ten newly collected Piper real-robot demonstrations outside the 1,100-case scored pool appear in Section~\ref{sec:real_robot_demonstrations}, Figs.~\ref{fig:real_robot_tasks_first}--\ref{fig:real_robot_tasks_second}). The single-arm simulation core (Group B, 300 episodes) comprises 120 PickCube, 80 PegInsertionSide, 40 PlugCharger, and 60 StackCube episodes. Cosmos3 pairs this core with 100 real-world episodes (5 apple-to-bowl, 35 object-to-box, 40 marker-insertion, and 20 block-stacking), whereas GR00T pairs it with 100 additional simulated episodes (40 PickCube, 40 PlugCharger, and 20 StackCube).

\subsection{Constructed Simulation References}
Following Section~\ref{sec:agentic_harness}, the constructed pool provides 200 embodiment-specific canonical references: 100 dual-arm ARX5 episodes and 100 single-arm Panda episodes. Each cohort spans five household manipulation families with 20 episodes per family: cabinet articulation (10 doors and 10 drawers), cup uprighting (covering overturned cups and cylindrical bottles, including the bottle-uprighting development checks in Fig.~\ref{fig:revisions}), cup hanging, tool use, and tissue-box pushing.

During synthesis, task-specific expert controllers combine inverse kinematics with collision-checked joint-space planning (direct interpolation or RRT-Connect) over SceneCode~\cite{wang2026scenecode} layouts and parametric primitives in PyBullet~\cite{coumans2016pybullet}. Automated goal predicates then verify joint progress, target pose, physical contact, post-release static support, hanging stability, or surface displacement; for instance, the uprighting predicate requires the object's upright axis to tilt by at most $12^\circ$ from vertical. Only trajectories passing all checks are replayed and rendered in Blender or SAPIEN~\cite{xiang2020sapien} to form versioned canonical references. For GR00T, public episodes run in ManiSkill, whereas constructed cases run PyBullet physics with live SAPIEN RGB rendering (Table~\ref{tab:native_interfaces}).

\subsection{Native Model Interfaces and Evaluation Setup}
\label{sec:supp_native_conversion}

\paragraph{Reference-to-native format conversion.}
Each constructed canonical reference packages SceneCode SDF/glTF geometry, the PyBullet collision SDF or articulated URDF, and initial frame-zero link poses, joint states, and physical parameters (mass, inertia, friction, and restitution). Model-specific adapters translate this canonical package into each simulator's native format: SAPIEN adapters resolve glTF $Y$-up versus SDF $Z$-up axes, link-local transforms, mesh scales, and quaternion conventions, while Isaac-based adapters compile SDF/URDF assets into USD.

To place canonical scenes within each robot's reachable workspace, adapters apply a recorded rigid transform (yaw rotation and translation, plus optional $XY$ object-spacing compression on movable tabletop layouts). Before rollout inference, automated validation checks confirm that loaded poses and joint angles match the transformed reference within 2\,cm position, $5^\circ$ rotational/revolute, and 5\,mm prismatic tolerances, and verify that the target object is visible in all required camera views. Any loading or visibility error halts scoring as a pipeline failure rather than a model failure. This pre-inference workspace placement is strictly separate from CAE scoring, which applies no post-hoc cross-video alignment.

\paragraph{Conditioning inputs.}
Table~\ref{tab:native_interfaces} lists the native conditioning inputs and rollout modes for the 100 constructed episodes per model (public episodes retain their source camera setups). Dual-arm models receive one head view and two wrist views: Fast-WAM and Long-WAM condition on the initial 14-DoF joint/gripper state with the head view stacked above the wrist views, whereas LingBot-VA uses the reverse vertical stacking order without robot joint states. Long-WAM additionally takes a 48-step control history (four control steps per RGB frame, padded with the initial frame at reset), and both Fast-WAM and Long-WAM initialize Table30 episodes from fixed-aperture gripper estimates. Among single-arm models, Cosmos3 takes a single external RGB view and language instruction without joint states or history, whereas GR00T applies the DROID camera convention and receives live RGB and 8-DoF joint/gripper observations after each executed 8-step action chunk. Across all models, future reference frames, states, and actions are strictly excluded during inference.

\begin{table}[!ht]
\centering\small
\setlength{\tabcolsep}{4pt}
\renewcommand{\arraystretch}{1.14}
\caption{\textbf{Native interfaces for the 100 constructed cases per model.} Public episodes retain source-specific camera profiles. Each row specifies the conditioning inputs and rollout mode used to produce $\widehat{V}$.}
\label{tab:native_interfaces}
\begin{tabular*}{\textwidth}{@{\extracolsep{\fill}}>{\raggedright\arraybackslash}p{.13\textwidth}>{\raggedright\arraybackslash}p{.40\textwidth}>{\raggedright\arraybackslash}p{.41\textwidth}@{}}
\toprule
\suppTableHead{Model} & \suppTableHead{Conditioning inputs} & \suppTableHead{Rollout mode and frame rate}\\
\midrule
Fast-WAM & Head and two wrist RGB views, language goal, initial 14-DoF joint/gripper state & Autoregressive video--action rollout using self-generated RGB and predicted joint states, with RGB at 12.5\,Hz.\\
Long-WAM & Three RGB views, language goal, initial 14-DoF state, 48-step observation history & Joint video--action rollout with latent memory continuation over self-generated RGB and states, with RGB at 12.5\,Hz.\\
LingBot-VA & Three initial RGB views and language goal without robot joint state & Open-loop video--action generation without environment state feedback, with RGB at 10\,Hz.\\
Cosmos3 & Single external RGB view and language goal, state-free with zero history & Autoregressive DROID video generation conditioned on self-generated RGB frames, with RGB at 15\,Hz.\\
GR00T & Live external/wrist RGB views, language goal, measured 7-joint + 1-gripper state & Closed-loop execution of 8-step action chunks at 15\,Hz in simulation with RGB recorded at 20\,Hz.\\
\bottomrule
\end{tabular*}
\end{table}

\paragraph{Evaluator settings and media clock.}
\label{sec:supp_figure_sources}
Automated visual judgments use \texttt{Qwen3-VL-8B-Instruct} served via vLLM~0.11.2 in BF16 without quantization under greedy decoding (temperature $0$, top-$p$ $1$, seed $0$, max $1{,}024$ output tokens), while harness evolution sessions use \texttt{gpt-6.1-sol}. All videos retain their native frame rates and media clocks without temporal resampling. In Fig.~\ref{fig:evaluation_landscape}, bowl-stacking and avocado-placement examples show Fast-WAM rollouts, while block-stacking and banana-transfer examples show Cosmos3 rollouts (the single-view banana illustration is excluded from benchmark aggregates).

\section{Inherited WorldOlympiad Track Definitions}
\label{sec:preliminaries}

Following WorldOlympiad~\cite{zhao2026worldolympiad}, rollout videos $\widehat{V}$ are compared against reference videos $V$ across three tracks: \emph{Physical Faithfulness} ($P$), \emph{Geometric Consistency}, and \emph{Interaction Fidelity}. Subscript $\mathrm{base}$ denotes the inherited WorldOlympiad scores prior to adding CAE.

\paragraph{Physical faithfulness ($P$).}
Across 14 candidate physical rules grouped into mechanics, thermodynamics, and material properties (Table~\ref{tab:metric_hierarchy}), an MLLM first determines which rules are triggered in the reference video. For each triggered rule $q$ in an active group $g$, a visual judge compares $\widehat{V}$ with $V$ to output binary compliance $c_{g,q} \in \{0,1\}$ and confidence $w_{g,q} \in [0,1]$. Averaging rule scores $c_{g,q} w_{g,q}$ across the $N_g$ triggered rules in each active group and then across all $M$ active groups gives
\begin{equation}
P = \frac{1}{M}\sum_{g=1}^{M}\left(\frac{1}{N_g}\sum_{q=1}^{N_g} c_{g,q} w_{g,q}\right).
\label{eq:prelim_physical}
\end{equation}
If no physical rule is triggered in a case ($M = 0$), $P$ and $S_{\mathrm{all}}$ are marked N/A.

\paragraph{Inherited geometric consistency ($G_{\mathrm{base}}$).}
The inherited geometric score averages three equal-weighted spatial criteria---3D scene reconstruction ($G_{\mathrm{rec}}$), novel-view consistency ($G_{\mathrm{view}}$), and camera-trajectory agreement ($G_{\mathrm{cam}}$):
\begin{equation*}
G_{\mathrm{base}} = \tfrac{1}{3}\bigl(G_{\mathrm{rec}} + G_{\mathrm{view}} + G_{\mathrm{cam}}\bigr).
\end{equation*}

\paragraph{Inherited interaction fidelity ($I_{\mathrm{base}}$).}
\label{eq:interaction_judgments}
Across $K$ captioned video chunks, an MLLM rates local interaction within each chunk, inter-chunk transition smoothness (defined when $K>1$), and global task completion on a $[0,5]$ scale; dividing by $5$ normalizes each rating to $[0,1]$, yielding $I_{\mathrm{local}}$, $I_{\mathrm{trans}}$, and $I_{\mathrm{global}}$. Global is the overall rating assigned to each rollout. The judge also rates long-range temporal consistency (Long Range) and alignment with the task instruction (Global Text) separately. In parallel, semantic alignment is measured with CLIP~\cite{radford2021learning}. Frame--caption cosine similarity $S_{\mathrm{clip}}$ is linearly calibrated to $[0,1]$ using WorldOlympiad's fixed cosine bounds $[0.20, 0.40]$:
\begin{equation}
I_{\mathrm{clip}} = \operatorname{clip}\!\left(\frac{S_{\mathrm{clip}} - 0.20}{0.40 - 0.20},\, 0,\, 1\right).
\label{eq:clip_calibration}
\end{equation}
When all four terms apply ($K>1$), WorldOlympiad combines the three MLLM judgments ($0.30$ weight each) with $I_{\mathrm{clip}}$ ($0.10$ weight):
\begin{equation*}
I_{\mathrm{base}} = 0.30\bigl(I_{\mathrm{local}} + I_{\mathrm{trans}} + I_{\mathrm{global}}\bigr) + 0.10\,I_{\mathrm{clip}}.
\end{equation*}

\section{Metric Implementation and CAE Analysis}
\label{sec:supp_metrics}

Table~\ref{tab:metric_hierarchy} summarizes all sub-metrics across the three tracks, highlighting our two CAE additions in bold.

\subsection{Metric Applicability Across Episodes}
\label{sec:supp_applicability}
Table~\ref{tab:applicable_counts} lists applicable episode counts for metrics that depend on event occurrence; all geometric criteria, global/local interaction criteria, and both CAE scores apply to all 500 cases per model. Across all 2,500 evaluations, 2,486 runs trigger at least one physical rule and 14 constructed runs have no triggered physical rule (marked N/A on $P$ and All, rather than assigned zero). Minor 1-episode differences on the shared dual-arm set reflect independent MLLM rule-relevance caching per evaluation workspace under BF16 batch kernel reduction.

\begin{table}[!ht]
\centering\footnotesize
\renewcommand{\arraystretch}{1.10}
\setlength{\tabcolsep}{4pt}
\caption{\textbf{Applicable episode counts by subset and model.} The two subsets sum to the 500-case totals. Thermodynamics, color mixing, solubility, and combustibility have no applicable episodes. Trans.\ applies to multi-chunk episodes.}
\label{tab:applicable_counts}
\begin{tabular*}{\textwidth}{@{\extracolsep{\fill}}llccccccc@{}}
\toprule
\suppTableHead{Subset} & \suppTableHead{Model} & \suppTableHead{Physical / All} & \suppTableHead{Mech. / Grav.} & \suppTableHead{Buoy.} & \suppTableHead{Comp.} & \suppTableHead{Impact} & \suppTableHead{Hard.} & \suppTableHead{Trans.}\\
\midrule
\multirow{5}{*}{Public} & Fast-WAM & 400 & 400 & 1 & 11 & 162 & 241 & 301\\
 & Long-WAM & 400 & 400 & 1 & 11 & 162 & 242 & 301\\
 & LingBot-VA & 400 & 400 & 1 & 10 & 163 & 242 & 301\\
 & Cosmos3 & 400 & 400 & \textemdash & \textemdash & 79 & 350 & 153\\
 & GR00T & 400 & 400 & \textemdash & \textemdash & 41 & 351 & 152\\
\midrule
\multirow{5}{*}{Constructed} & Fast-WAM & 96 & 89 & \textemdash & 20 & 26 & 53 & 100\\
 & Long-WAM & 95 & 89 & \textemdash & 20 & 28 & 51 & 100\\
 & LingBot-VA & 95 & 90 & \textemdash & 20 & 28 & 50 & 100\\
 & Cosmos3 & 100 & 94 & \textemdash & 5 & 43 & 75 & 100\\
 & GR00T & 100 & 93 & \textemdash & 4 & 46 & 76 & 100\\
\bottomrule
\end{tabular*}
\end{table}

\subsection{Track Integration and Weight Renormalization}
\label{sec:supp_track_integration}
\label{eq:supp_applicability_aggregation}
\label{eq:overall_score}
WorldOlympiad 2.0 keeps $P$ unchanged and adds $S_{\mathrm{path}}$ to $G$ ($G = 0.75\,G_{\mathrm{base}} + 0.25\,S_{\mathrm{path}}$; Eq.~\eqref{eq:geometry}) and $S_{\mathrm{vel}}$ to $I$ (Eq.~\eqref{eq:interaction} when all five criteria apply). Whenever a criterion is inapplicable on a given episode, its weight is omitted and the remaining active weights are renormalized to sum to $1$ on that episode. In particular, on single-chunk episodes ($K=1$, where $I_{\mathrm{trans}}$ is N/A; counts in Table~\ref{tab:applicable_counts}), the remaining interaction weights $(0.24, 0.24, 0.08, 0.20)$ for $(I_{\mathrm{local}}, I_{\mathrm{global}}, I_{\mathrm{clip}},S_{\mathrm{vel}})$ are divided by $0.76$ per case before averaging across the 500 episodes---explaining why dataset-level means in Table~\ref{tab:interaction_model_details} differ slightly from applying $0.80$ and $0.20$ directly to the 500-episode column means of $I_{\mathrm{base}}$ and $S_{\mathrm{vel}}$. For all cases with a valid physical score, the overall score is $S_{\mathrm{all}} = (P + G + I)/3$.

\begin{table}[!ht]
\centering\small
\setlength{\tabcolsep}{4pt}
\renewcommand{\arraystretch}{1.10}
\caption{\textbf{Evaluation metric hierarchy.} Bold entries denote CAE additions introduced in WorldOlympiad 2.0. The remaining criteria are inherited from WorldOlympiad~\cite{zhao2026worldolympiad}.}
\label{tab:metric_hierarchy}
\begin{tabular*}{\textwidth}{@{\extracolsep{\fill}}>{\raggedright\arraybackslash}p{.19\textwidth}>{\raggedright\arraybackslash}p{.21\textwidth}>{\raggedright\arraybackslash}p{.54\textwidth}@{}}
\toprule
\suppTableHead{Track} & \suppTableHead{Metric / group} & \suppTableHead{Evaluated criteria}\\
\midrule
Physical faithfulness & Mechanics & Gravity, buoyancy, compression, impact\\
 & Thermodynamics & Melting, sublimation, vaporization, condensation, deposition, freezing\\
 & Material properties & Color mixing, solubility, hardness, combustibility\\
\midrule
Geometric consistency & Scene reconstruction & Object layout, spatial coherence, cross-view stability, scene alignment\\
 & Meta-view consistency & Novel-view rendering consistency at the farthest recovered camera pose\\
 & Camera trajectory & Camera path shape, translation scale/center, rotation geodesic error\\
 & \textbf{CAE action path} & Speed-invariant 3D object path fidelity over normalized arc-length progress\\
\midrule
Interaction fidelity & Local interaction & Per-chunk visual quality, instruction alignment, overall chunk judgment\\
 & Transition & Inter-chunk boundary smoothness and state continuity for $K>1$\\
 & Global & Overall rollout rating, with separate ratings of long-range consistency and instruction alignment\\
 & Semantic alignment & Frame--caption cosine similarity calibrated to $[0,1]$ using Eq.~\eqref{eq:clip_calibration}\\
 & \textbf{CAE velocity vectors} & Physical-time 3D velocity direction and speed fidelity over $[0,T]$\\
\bottomrule
\end{tabular*}
\end{table}

\subsection{Continuous Action Trajectory Recovery from RGB Videos}
\label{sec:supp_trajectory_extraction}

\paragraph{Depth scale calibration and rigid pose fitting.}
\label{eq:supp_rigid_solution}
\label{eq:supp_pose_interpolation}
Following Dream2Flow~\cite{dharmarajan2026dream2flow}, we recover 3D object flow separately in $V$ and $\widehat{V}$ on the primary fixed external (or head) RGB view so wrist-camera ego-motion does not confound object trajectories. Specifically, SAM~3~\cite{carion2026sam} segments the target object from the task prompt, CoTracker3~\cite{karaev2025cotracker3} tracks $16$--$64$ foreground query points across all native video frames, and Depth Anything~3~\cite{lin2025depth} predicts per-frame monocular depth to lift visible 2D tracks into 3D. Applying this identical perception operator symmetrically to $V$ and $\widehat{V}$ compares reference and rollout motion in a shared visual observation space without requiring ground-truth simulator object states, which video-only world models do not produce. Reference frame-zero camera intrinsics and metric depth define a shared coordinate frame; each video's predicted depth sequence is scaled once by least squares against frame-zero depth on tracked target pixels and kept fixed across all subsequent frames, using no simulator states, future reference depth, or action outputs.

Because partial occlusions during grasping can shift the mean of visible particles, we fit a $3\times 3$ rigid rotation matrix $R_i$ and 3D translation $\mathbf{b}_i$ at each frame $i$ relative to the first valid anchor frame $a$ (having at least three tracked 3D particles with centroid $\mathbf{c}_a$) to obtain the frame-$i$ position on the action trajectory $\mathbf{x}_i$:
\begin{equation}
(R_i,\mathbf{b}_i)=\underset{R,\,\mathbf{b}}{\arg\min}
\sum_{j\in J_i}\|R\mathbf{p}_{a,j}+\mathbf{b}-\mathbf{p}_{i,j}\|^2,\qquad
\mathbf{x}_i = R_i\mathbf{c}_a + \mathbf{b}_i,
\label{eq:supp_rigid_recovery}
\end{equation}
where $\mathbf{p}_{i,j}$ is particle $j$'s 3D position at frame $i$, $J_i$ is the set of particles co-visible in frames $a$ and $i$ ($|J_i|\geq 3$), and $(R_i,\mathbf{b}_i)$ is solved in closed form via SVD rigid alignment. Occluded frames with $|J_i|<3$ are skipped during pose fitting; the continuous action trajectory linearly interpolates valid positions across native timestamps $t_i = i/f$ at frame rate $f$ (clamped at boundary poses outside the first/last valid detections, or kept constant at $\mathbf{x}_0$ if fewer than three particles are tracked throughout) over $[0,T]$, without post-hoc cross-video spatial or temporal registration.

\paragraph{Cross-rate bowl-stacking example.}
In Fig.~\ref{fig:ee_continuous_motion} (Table30 \texttt{stack\_bowls}, episode 1), CAE compares a 400-frame (30\,Hz) real-robot reference directly against LingBot-VA's 134-frame (10\,Hz) rollout over $[0, 13.3]$\,s without frame resampling, yielding $S_{\mathrm{path}}=0.603$ and $S_{\mathrm{vel}}=0.297$ (with direct particle-tracking support on $87.5\%$ and $99.3\%$ of frames, respectively).

\subsection{Implementation of Inherited Visual Judges}
\label{sec:physical_faithfulness}
For Physical Faithfulness ($P$), an MLLM identifies moving or deforming objects and SAM~3~\cite{carion2026sam} extracts object-centric crops for rule-level compliance scoring (Eq.~\eqref{eq:prelim_physical}). For inherited Geometric Consistency ($G_{\mathrm{base}}$), Depth Anything~3~\cite{lin2025depth} estimates scene geometry and camera poses from up to 32 uniformly sampled frames, and 3D Gaussian Splatting~\cite{kerbl20233d} renders reconstructed and farthest-viewpoint images scored by visual judges to obtain $G_{\mathrm{rec}}$ and $G_{\mathrm{view}}$, while $G_{\mathrm{cam}}$ measures similarity-aligned camera-trajectory agreement. Unlike $G_{\mathrm{cam}}$, CAE applies no similarity alignment between reference and rollout action trajectories so that true spatial offsets and scale errors are penalized.

\subsection{Closed-Form Path and Velocity Integration}
\label{sec:geometric_consistency}
\label{sec:interaction_fidelity}
\label{eq:cae_similarity}

\paragraph{Exact path integration.}
Given recovered 3D positions $\mathbf{x}_0,\ldots,\mathbf{x}_n$ across $n+1$ frames with segment lengths $d_i=\|\mathbf{x}_{i+1}-\mathbf{x}_i\|$ ($i=0,\ldots,n-1$) and total path length $L=\sum_{i=0}^{n-1} d_i > 0$, normalized arc-length grid points are $s_0=0$ and $s_i = L^{-1}\sum_{j<i} d_j$ (zero-length pause segments collapse to a single point in $s$ while retaining their duration in physical time $t$; if $L=0$, the path is constant). Merging the reference and rollout progress grids into a sorted grid $0 = s_0 < \cdots < s_M = 1$ of $M$ sub-intervals makes $\mathbf{q}(s)$, $\widehat{\mathbf{q}}(s)$, and $\mathbf{q}(s)-\widehat{\mathbf{q}}(s)$ linear on every sub-interval $[s_m, s_{m+1}]$. For any piecewise-linear 3D signal $\mathbf{z}(s)$ with grid values $\mathbf{z}_m = \mathbf{z}(s_m)$, its integrated squared norm on $[0,1]$ evaluates in exact closed form:
\begin{equation}
\|\mathbf{z}\|_{[0,1]}^2
=\sum_{m=0}^{M-1}\frac{s_{m+1}-s_m}{3}
\bigl(\|\mathbf{z}_m\|^2+\mathbf{z}_m^\top\mathbf{z}_{m+1}+\|\mathbf{z}_{m+1}\|^2\bigr),
\label{eq:supp_exact_path_energy}
\end{equation}
which directly computes $\|\mathbf{q}-\widehat{\mathbf{q}}\|_{[0,1]}$, $\|\mathbf{q}\|_{[0,1]}$, and $\|\widehat{\mathbf{q}}\|_{[0,1]}$ in Eq.~\eqref{eq:path}.

\paragraph{Exact velocity integration.}
Similarly, piecewise-linear trajectories $\mathbf{x}(t)$ and $\widehat{\mathbf{x}}(t)$ have constant velocities on each native frame interval. Merging both videos' frame timestamps on $[0,T]$ into a sorted grid $0 = \tau_0 < \cdots < \tau_J = T$ of $J$ sub-intervals yields constant reference and rollout velocities $\mathbf{v}_j, \widehat{\mathbf{v}}_j$ on each sub-interval $\Delta\tau_j = \tau_{j+1}-\tau_j$, giving exact squared norms
\begin{equation}
\begin{aligned}
\|\mathbf{v}\|_{[0,T]}^2 &= \sum_{j=0}^{J-1}\Delta\tau_j\|\mathbf{v}_j\|^2,\quad
\|\widehat{\mathbf{v}}\|_{[0,T]}^2 = \sum_{j=0}^{J-1}\Delta\tau_j\|\widehat{\mathbf{v}}_j\|^2,\\
\|\mathbf{v}-\widehat{\mathbf{v}}\|_{[0,T]}^2 &= \sum_{j=0}^{J-1}\Delta\tau_j\|\mathbf{v}_j-\widehat{\mathbf{v}}_j\|^2
\end{aligned}
\label{eq:supp_exact_velocity_energy}
\end{equation}
for $S_{\mathrm{vel}}$ in Eq.~\eqref{eq:velocity}, without resampling either video to a common frame rate.

\subsection{Theoretical Properties of CAE}
\label{sec:supp_cae_properties}

\paragraph{Speed-magnitude and direction decomposition.}
At any time $t$ with speeds $v=\|\mathbf{v}(t)\|$, $\widehat{v}=\|\widehat{\mathbf{v}}(t)\|$, and angle $\theta$ between the two velocity vectors, the squared velocity difference decomposes into speed mismatch and directional error:
\begin{equation}
\|\mathbf{v}-\widehat{\mathbf{v}}\|^2 = (v-\widehat{v})^2 + 2v\widehat{v}(1-\cos\theta),
\label{eq:supp_velocity_decomposition}
\end{equation}
so opposite-direction motion ($\theta=\pi$) at equal speed ($v=\widehat{v}>0$) yields $\|\mathbf{v}-\widehat{\mathbf{v}}\| = 2v = \|\mathbf{v}\| + \|\widehat{\mathbf{v}}\|$ and scores $S_{\mathrm{vel}} = 0$.

\paragraph{Speed-profile invariance and joint uniqueness.}
\textbf{Proposition.} \emph{Let $\mathbf{x}(t)$ be a moving piecewise-linear trajectory on $[0,T]$ and $\widehat{\mathbf{x}}(t)=\mathbf{x}(\phi(t))$ for any continuous, nondecreasing, piecewise-linear time-warping function $\phi:[0,T]\to[0,T]$ with $\phi(0)=0$ and $\phi(T)=T$. Then $S_{\mathrm{path}}=1$, and $S_{\mathrm{vel}}<1$ if the velocity vectors differ over a nonzero time interval. Moreover, $S_{\mathrm{path}}=S_{\mathrm{vel}}=1$ holds if and only if $\widehat{\mathbf{x}}(t)=\mathbf{x}(t)$ for all $t\in[0,T]$.}

\emph{Proof.} By the chain rule, $\widehat{a}(t)=\int_0^t\|\dot{\mathbf{x}}(\phi(u))\|\phi'(u)\,du = a(\phi(t))$, so $\widehat{L}=L$ and both trajectories pass through the exact same 3D points at every travel progress $s\in[0,1]$:
\begin{equation}
\widehat{a}(t)=a(\phi(t)),\qquad \widehat{L}=L,\qquad \widehat{\mathbf{v}}(t)=\mathbf{v}(\phi(t))\,\phi'(t).
\label{eq:supp_pace_invariance}
\end{equation}
Since $\widehat{\mathbf{x}}(0)=\mathbf{x}(0)$, start anchoring gives $\widehat{\mathbf{q}}(s)=\mathbf{q}(s)$ and $S_{\mathrm{path}}=1$. A time change confined to a pause leaves the velocity unchanged. A velocity difference over a nonzero interval gives $\|\widehat{\mathbf{v}}-\mathbf{v}\|_{[0,T]}>0$, hence $S_{\mathrm{vel}}<1$. Conversely, $S_{\mathrm{vel}}=1$ implies $\widehat{\mathbf{v}}(t)=\mathbf{v}(t)$ almost everywhere, so $\widehat{\mathbf{x}}(t)-\mathbf{x}(t)=\boldsymbol{\delta}$ for a constant offset $\boldsymbol{\delta}$; start anchoring at $\mathbf{x}(0)$ then gives $\widehat{\mathbf{q}}(s)-\mathbf{q}(s)=\boldsymbol{\delta}$, and $S_{\mathrm{path}}=1$ forces $\boldsymbol{\delta}=\mathbf{0}$, i.e., $\widehat{\mathbf{x}}(t)=\mathbf{x}(t)$ across $[0,T]$. $\square$

\paragraph{Similarity invariance.}
Applying a shared global similarity transform $\mathbf{x}\mapsto \lambda R\mathbf{x}+\mathbf{b}$ (scale $\lambda>0$, $3\times 3$ rotation matrix $R$, translation $\mathbf{b}\in\mathbb{R}^3$) or uniform clock scaling $t'=ct$ ($c>0$) to both reference and rollout cancels $\mathbf{b}$ via start anchoring and scales numerator and denominator norms identically, leaving both $S_{\mathrm{path}}$ and $S_{\mathrm{vel}}$ invariant.

\subsection{Controlled Trajectory Validation}
\label{sec:supp_controls}
To verify these analytical properties numerically, we apply controlled spatial and temporal perturbations to a 201-sample reference curve on $t\in[0,10]$\,s with normalized time $u=t/10 \in [0,1]$,\allowbreak\ $\mathbf{x}(t)=[0.25u,\, 0.10\sin(\pi u),\, 0.03\sin(2\pi u)]^\top$ (Table~\ref{tab:controlled}).

\begin{table}[!ht]
\centering\small
\begin{minipage}{0.74\textwidth}
\centering
\renewcommand{\arraystretch}{1.10}
\setlength{\tabcolsep}{8pt}
\caption{\textbf{Controlled trajectory perturbations under CAE.} Spatial and temporal perturbations confirm orthogonal sensitivity of $S_{\mathrm{path}}$ and $S_{\mathrm{vel}}$. Path and Vel.\ report $S_{\mathrm{path}}$ and $S_{\mathrm{vel}}$. Sampled polylines match $1.000$ to five decimal places.}
\label{tab:controlled}
\begin{tabular*}{\linewidth}{@{\extracolsep{\fill}}lrr@{}}
\toprule
\suppTableHead{Perturbation} & \suppTableHead{Path} & \suppTableHead{Vel.}\\
\midrule
Identical trajectory & 1.000 & 1.000\\
Same path, nonlinear speed $u^2$ & 1.000 & 0.625\\
Wait, then move & 1.000 & 0.677\\
Constant 5\,cm spatial offset & 0.857 & 1.000\\
Half trajectory extent, $c=0.5$ & 0.667 & 0.667\\
Opposite direction & 0.000 & 0.000\\
Stationary rollout & 0.000 & 0.000\\
Mid-rollout reversal & 0.632 & 0.304\\
\bottomrule
\end{tabular*}

\end{minipage}
\end{table}

Nonlinear pacing ($u^2$) and an initial 2\,s pause preserve $S_{\mathrm{path}}=1.000$ while lowering $S_{\mathrm{vel}}$ to $0.625$ and $0.677$. Conversely, a constant 5\,cm spatial shift preserves $S_{\mathrm{vel}}=1.000$ while reducing $S_{\mathrm{path}}$ to $0.857$. Scaling trajectory amplitude by $c=0.5$ yields $0.667$ ($2/3$) on both metrics, matching the exact identity for scaled signals ($c\geq 0, f\neq 0$):
\begin{equation}
\mathcal{S}_D(f, cf) = 1 - \frac{|1-c|}{1+c} = \frac{2\min(1,c)}{1+c}.
\label{eq:supp_proportional_control}
\end{equation}
Opposite-direction and stationary rollouts score $0.000$, and mid-rollout reversal penalizes both metrics ($0.632$ and $0.304$). These synthetic checks are used solely for metric validation and are excluded from the 2,500 benchmark evaluations.

\subsection{Scope and Limitations}
\label{sec:limitations}
Because CAE recovers continuous action trajectories from RGB videos, its precision depends on foreground segmentation, point tracking, monocular depth estimation, and target visibility. Recovered coordinates are expressed in the initial camera frame (retaining camera motion if the viewpoint moes), and tracking the target centroid does not capture in-place object rotation, joint-limit violations, or contact forces. In addition, reference-anchored trajectory matching can penalize alternative valid manipulation routes that achieve the same task goal via a different path, which thus complements rather than replaces semantic goal judgments. Finally, single-arm scores reflect each model's native conditioning interface and episode split rather than a shared training-disjoint leaderboard.

\section{Agent Harness Implementation and Algorithms}
\label{sec:supp_harness}
Our evaluation harness coordinates reference synthesis, native adaptation, rollout scoring, and closed-loop repair through stage checkpoints and persistent diagnostic records (logs of simulator conventions, interface specifications, and verified repairs). As shown in Fig.~\ref{fig:method_overview}, reusable procedural routines are organized into six workflow groups---\emph{Observe \& Orchestrate}, \emph{Task \& Asset}, \emph{Simulation \& Reference}, \emph{Interface \& Camera}, \emph{Measure \& Evaluate}, and \emph{Revise \& Remember} (workflow categories rather than newly learned skills). In each round, the agent inspects checkpoints to schedule pending case--model pairs while leaving active jobs untouched. Whenever a stage fails validation, candidate edits to tool scripts, workflow instructions, or persistent records are tested in a sandbox against focused regression checks before promotion, ensuring in-flight jobs retain a locked harness version.

Any repair that modifies a constructed scene or goal check increments the reference version and triggers full revalidation and re-rendering, binding every model score to an immutable canonical reference (Algorithms~\ref{alg:synthesis} and~\ref{alg:evaluation}). Across the frozen benchmark manifest, all 2,500 scheduled model--case evaluations (2,000 public and 500 constructed, drawn from 900 public and 200 constructed references) complete under locked scoring rules.

\subsection{Reference Construction and Evaluation Algorithms}
\label{sec:supp_algorithms}
In Section~\ref{sec:harness_results}, we count every repair attempt---including retries on the same fault---and average across cases in each group, while the recovery rate denotes the fraction of recorded pipeline failures resolved prior to scoring ($24$ vs.\ $0.49$ mean attempts per constructed vs.\ public case, with $100\%$ of recorded failures resolved). Structurally, constructed cases incur heavier repair activity because they traverse both Algorithm~\ref{alg:synthesis} (synthesizing SceneCode geometry and collision meshes, tuning IK/RRT-Connect contact trajectories, and passing physical post-release predicates in PyBullet) and Algorithm~\ref{alg:evaluation} (cross-simulator adaptation), whereas curated public episodes enter Algorithm~\ref{alg:evaluation} directly with pre-validated demonstrations.

\begin{algorithm}[!ht]
\caption{Physics-checked canonical reference synthesis}
\label{alg:synthesis}
\small
\algrenewcommand\algorithmicindent{1.4em}
\begin{algorithmic}[1]
\Require Task spec $s_{\mathrm{task}}$, harness $H$, diagnostic log $e$, max rounds $N$
\Ensure Canonical reference $(V,\text{meta})$, updated $H$ and $e$
\For{$r=1,\ldots,N$}
    \State $(\text{scene}, e) \gets \textsc{GenerateScene}(s_{\mathrm{task}}, H, e)$
    \State $(\text{sim}, e) \gets \textsc{BuildPyBullet}(\text{scene}, s_{\mathrm{task}}, e)$
    \State $(\text{traj}, e) \gets \textsc{RunExpertPolicy}(\text{sim}, s_{\mathrm{task}}, e)$
    \If{$\textsc{CheckGoalPredicates}(\text{traj}, s_{\mathrm{task}}, e)$}
        \State $(V, e) \gets \textsc{RenderMultiView}(\text{scene}, \text{traj}, s_{\mathrm{task}}, e)$
        \If{$\textsc{VerifyReference}(s_{\mathrm{task}}, \text{scene}, \text{traj}, V, e)$}
            \State $\text{meta} \gets (s_{\mathrm{task}}, \operatorname{version}(\text{scene}), V)$
            \State \Return $(V, \text{meta}, H, e)$
        \EndIf
    \EndIf
    \State $(\text{stage}, \text{patch}) \gets \textsc{DiagnoseAndPatch}(H, e)$ \label{ln:synth_diagnose}
    \If{$\textsc{PassRegression}(\textsc{Apply}(H, \text{patch}), e)$}
        \State $H \gets \textsc{Apply}(H, \text{patch})$
        \State $e \gets \textsc{ResetFromStage}(H, \text{stage}, e)$ \label{ln:synth_retry}
    \EndIf
\EndFor
\State \Return $\text{failure}$
\end{algorithmic}
\end{algorithm}

\begin{algorithm}[!ht]
\caption{Native adaptation, rollout scoring, and repair}
\label{alg:evaluation}
\small
\algrenewcommand\algorithmicindent{1.4em}
\begin{algorithmic}[1]
\Require Queue $Q$ of $(\text{case }c, \text{model }m)$ pairs, references $\text{Refs}$, harness $H$, max rounds $N$
\Ensure Summary report $R$, results $\text{Res}$, updated $H$
\State $\text{Res} \gets \{\}$;\; initialize logs $E \gets \{e_i : i \in Q\}$
\For{$r=1,\ldots,N$}
    \For{each $i=(c,m) \in \textsc{Schedule}(Q, \text{Res}, E)$}
        \State $(V, \text{meta}) \gets \text{Refs}[c]$
        \State $(\text{pkg}_i, e_i) \gets \textsc{BuildNativePkg}(c, m, \text{meta}, H, e_i)$
        \If{$\textsc{ValidPackage}(\text{pkg}_i, e_i)$}
            \State $(\widehat{V}, e_i) \gets \textsc{RunModelRollout}(m, \text{pkg}_i, e_i)$
            \State $(\mathbf{x}, \widehat{\mathbf{x}}, e_i) \gets \textsc{RecoverTraj}(V, \widehat{V}, c, e_i)$
            \State $\text{scores}_i \gets \textsc{ScoreAllTracks}(V, \widehat{V}, \mathbf{x}, \widehat{\mathbf{x}}, e_i)$
        \EndIf
        \If{$\textsc{PipelineError}(e_i)$}
            \State $(\text{stage}, \text{patch}) \gets \textsc{DiagnoseAndPatch}(H, e_i)$
            \If{$\textsc{PassRegression}(\textsc{Apply}(H, \text{patch}), e_i)$}
                \State $H \gets \textsc{Apply}(H, \text{patch})$
                \State $e_i \gets \textsc{ResetFromStage}(H, \text{stage}, e_i)$
            \EndIf
        \Else
            \State $\text{Res}[i] \gets (\text{scores}_i, \text{meta}, e_i)$
        \EndIf
    \EndFor
    \State \textbf{if} $|\text{Res}| = |Q|$ \textbf{then return} $(\textsc{Summarize}(\text{Res}), \text{Res}, H)$
\EndFor
\State \Return $\text{incomplete}(\text{Res}, H)$
\end{algorithmic}
\end{algorithm}

\subsection{Scoped Development Progress, Revision Reuse, and Execution Traces}
\label{sec:supp_harness_traces}
To illustrate how Algorithms~\ref{alg:synthesis} and~\ref{alg:evaluation} operate in practice, we summarize archived development preflight records, traceable revision promotions, and execution-stage receipts below. These retrospective, scoped records document concrete repair and reuse episodes; they do not estimate a benchmark-wide first-pass rate or a controlled causal ablation of disabling persistent records.

\paragraph{Fixed-cohort physical preflight progression.}
Across three tracked development cohorts of constructed references, physics-only preflight checks (prior to multi-view rendering and final admission) progressed from partial or zero initial acceptance to full cohort completion:

\begin{center}
\small
\setlength{\tabcolsep}{12pt}
\renewcommand{\arraystretch}{1.12}
\begin{tabular}{@{}lccc@{}}
\toprule
\suppTableHead{Development cohort} & $N$ & \suppTableHead{Initial} & \suppTableHead{After repair} \\
\midrule
Cabinet-door articulation & 10 & $0/10$, $0\%$ & $10/10$, $100\%$ \\
Cup uprighting & 20 & $13/20$, $65\%$ & $20/20$, $100\%$ \\
Dual-arm tissue-box pushing & 20 & $7/20$, $35\%$ & $20/20$, $100\%$ \\
\bottomrule
\end{tabular}
\end{center}

On the fixed 10-case \emph{cabinet-door} cohort, cumulative preflight passes progressed from $0/10$ on the first controller trial (and $0/10$ at the end of revision \texttt{v1}) to $3/10$ by \texttt{v3}, $8/10$ by \texttt{v4}, $9/10$ by \texttt{v9}, and $10/10$ at the verified \texttt{v19} preflight. The first 16 version ledgers record $334$ controller trials in total, and the separate \texttt{v19} search on the final unresolved case (\texttt{cabinet\_door\_019}) records $8$ candidate trials ($6$ failing the goal predicate, including $2$ that simultaneously failed the 240\,Hz post-release no-contact and settling-drift checks, before passing on trials $7$--$8$ alongside the clearance audit). Because later versions focus on remaining unresolved cases, this series reflects cumulative completion on a fixed set rather than independent per-round trial averages.

On \emph{cup uprighting} ($20$ fixed case IDs under a shared plan hash), $13$ cases retained the initial controller while the $7$ recovered cases succeeded via a first fixed-order retry ($3$), a second retry ($1$), current-joint-rest IK ($1$), or current-joint-rest IK with an airborne approach shift ($2$). On \emph{dual-arm tissue-box pushing} ($20$ cases), $7$ passed under baseline controller \texttt{v3}, while $7$ first passed with \texttt{v6} and $6$ with \texttt{v7} (counts classify selected passing recipes rather than total attempts). Separately, a 20-case dual-arm cup-hanging pilot recorded $17/20$ initial physical preflight passes and $20/20$ completions on a distinct downstream adapter-input catch-up gate ($0$ listed failures).

\paragraph{Promotions, focused checks, and reuse.}
Archived receipts document concrete promotion and reuse cycles corresponding to Fig.~\ref{fig:revisions}:
(i)~In an engineering-assisted \emph{hanging asset} revision incorporating measured mug-and-rack geometry, task success on the asset-factory assessment rose from $0.0$ (baseline) to $1.0$ on both candidate and holdout gates with no flagged regressions; the promoted asset was then reused directly in a separate hanging task (\texttt{hang\_reuse\_d}), passing its execution certificate.
(ii)~For \emph{bottle uprighting}, two targeted revisions---fixing the declared $+Y$ semantic upright axis (\texttt{5cbfbc6}) and pinned-source asset loading (\texttt{e46d776})---passed two focused verification suites of $13$ axis checks and $11$ loading checks ($24$ reported checks across those two suites).
(iii)~A manually proposed, automatically promoted passive \emph{reference-camera} revision (\texttt{a22db3f}) improved camera-visibility check scores on development ($0.184$ to $0.240$, bottle), holdout ($0.459$ to $0.663$, kettle/coaster), and regression ($0.049$ to $0.070$, tool use) checks, and was subsequently reused on a later mug task (passing the camera check at $0.348$; camera-check scores are distinct from task or benchmark scores).

\paragraph{Native adaptation and execution-stage receipts.}
Diagnostic logs from Algorithm~\ref{alg:evaluation} likewise separate shared environment faults from model performance. In a 100-case Cosmos3 constructed diagnostic batch, $20$ generation runs completed initially while $80$ halted on one shared missing Python module; resolving that single dependency allowed all $80$ cases to complete on retry ($100/100$ generation completions prior to downstream admission). On archived LingBot-VA public-episode receipts ($276$ unique case IDs: $274$ terminal completions and $2$ intermediate snapshots), $1/274$ ($\approx 0.36\%$) recorded a transient CUDA out-of-memory error followed by completion on retry under the same source binding; separate control receipts log a Table30 frame-horizon mismatch ($133$ vs.\ $134$ required frames) and a two-case RoboMIND depth-support requeue during interface and scoring checks.

\section{Detailed Benchmark Breakdowns and Human Evaluation}
\label{sec:supp_track_details}

\subsection{Fine-Grained Interaction Sub-Scores}
\label{sec:supp_model_results}
Table~\ref{tab:interaction_criteria_details} reports fine-grained diagnostic interaction sub-scores averaged over each model's applicable episodes: local visual quality (Visual), local instruction alignment (Instr.), inter-chunk smoothness (Smooth.), and the combined MLLM interaction score (MLLM). These diagnostic criteria complement the local, transition, and global MLLM ratings reported in Table~\ref{tab:interaction_model_details}.

\subsection{Human Evaluation Protocol}
\label{sec:supp_human}
\label{eq:supp_human_mean}
Our human evaluation study received institutional ethics approval (reviewing body anonymized for double-blind review), and all ten human raters provided informed consent. Rollout videos were presented alongside the task instruction and canonical reference video $V$ in randomized, model-blinded order, and each rater independently scored all applicable episodes across the physical, geometric, and interaction tracks on $[0,1]$. For each episode and track, we average the ten raters' scores to obtain a human track rating, average the three tracks per episode when all three apply, and average across applicable episodes per model---matching our automatic aggregation protocol exactly---to compute Spearman rank correlation ($\rho=0.90$) in Table~\ref{tab:human_eval}. Notably, human raters place LingBot-VA ($0.608$) ahead of GR00T ($0.595$), whereas automatic overall scores invert the top two ($0.519$ vs.\ $0.643$): raters penalize GR00T's inter-chunk execution pauses and trajectory drift (reflected in its low CAE scores and $I_{\mathrm{trans}}=0.208$) more heavily than scene-level 3D reconstruction judges ($G_{\mathrm{cam}}=0.903$).

\subsection{Public vs.\ Constructed Breakdowns}
Tables~\ref{tab:cae_source_split} and~\ref{tab:curated_results} separate CAE scores and track-level results between each model's 400 public episodes and 100 constructed simulation episodes, whose applicable-case-weighted combination equals the 500-case means in Table~\ref{tab:main_results} (rule-, criterion-, and family-level decompositions with applicable counts follow in Sections~\ref{sec:subset_detailed_results}--\ref{sec:synthetic_component}). Three diagnostic patterns stand out. First, on the 400 shared dual-arm public episodes, frequent grasp misses leave target objects nearly stationary across models, compressing $S_{\mathrm{path}}$ to $0.250$--$0.261$, whereas constructed multi-stage tasks separate them ($0.171$--$0.214$), with Long-WAM ($0.214$) benefiting from its 48-step observation history. Second, state-free zero-history Cosmos3 drops sharply in $S_{\mathrm{path}}$ from public ($0.434$) to constructed ($0.174$) episodes on multi-step articulation and contact tasks, whereas closed-loop simulated GR00T preserves path fidelity ($0.237$ to $0.246$). Third, $S_{\mathrm{vel}}$ increases across all five models on constructed cases ($0.211$--$0.231$ vs.\ $0.127$--$0.173$ on public episodes) because physics-checked expert controllers follow uniform motion pacing, whereas public teleoperated demonstrations exhibit higher speed variance.

\begin{table}[!ht]
\centering
\begin{minipage}[t]{.48\textwidth}
\centering\small
\setlength{\tabcolsep}{3pt}\renewcommand{\arraystretch}{1.10}
\caption{\textbf{Fine-grained interaction sub-scores.}}
\label{tab:interaction_criteria_details}
\label{tab:interaction_fine_details}
\begin{tabular*}{\linewidth}{@{\extracolsep{\fill}}lrrrr@{}}
\toprule
\suppTableHead{Model} & \suppTableHead{Visual} & \suppTableHead{Instr.} & \suppTableHead{Smooth.} & \suppTableHead{MLLM}\\
\midrule
Fast-WAM & 0.385 & 0.343 & 0.244 & 0.368\\
Long-WAM & 0.429 & 0.376 & 0.369 & 0.444\\
LingBot-VA & 0.513 & 0.413 & 0.549 & 0.523\\
\midrule
Cosmos3 & 0.431 & 0.355 & 0.235 & 0.312\\
GR00T & 0.853 & 0.447 & 0.208 & 0.563\\
\bottomrule
\end{tabular*}
\end{minipage}\hfill
\begin{minipage}[t]{.48\textwidth}
\centering\small
\setlength{\tabcolsep}{3pt}\renewcommand{\arraystretch}{1.10}
\caption{\textbf{CAE scores by reference source.}}
\label{tab:cae_source_split}
\begin{tabular*}{\linewidth}{@{\extracolsep{\fill}}lrrrr@{}}
\toprule
\multirow{2}{*}{\suppTableHead{Model}} & \multicolumn{2}{c}{\suppTableHead{Public}} & \multicolumn{2}{c}{\suppTableHead{Constructed}}\\
\cmidrule(lr){2-3}\cmidrule(lr){4-5}
 & \suppTableHead{Path} & \suppTableHead{Vel.} & \suppTableHead{Path} & \suppTableHead{Vel.}\\
\midrule
Fast-WAM & 0.260 & 0.143 & 0.171 & 0.211\\
Long-WAM & 0.250 & 0.138 & 0.214 & 0.231\\
LingBot-VA & 0.261 & 0.144 & 0.184 & 0.218\\
\midrule
Cosmos3 & 0.434 & 0.173 & 0.174 & 0.220\\
GR00T & 0.237 & 0.127 & 0.246 & 0.224\\
\bottomrule
\end{tabular*}
\end{minipage}
\end{table}

\begin{table}[!ht]
\centering\small
\renewcommand{\arraystretch}{1.08}
\setlength{\tabcolsep}{4pt}
\caption{\textbf{Track scores on public and constructed subsets.} All is the per-case average of Physical, 3D Cons., and Interact. Table~\ref{tab:applicable_counts} gives applicable counts.}
\label{tab:curated_results}
\label{tab:synthetic_results}
\begin{tabular*}{\textwidth}{@{\extracolsep{\fill}}llrrrrrr@{}}
\toprule
\suppTableHead{Subset} & \suppTableHead{Model} & \suppTableHead{Physical} & \suppTableHead{3D Cons.} & \suppTableHead{Interact.} & \suppTableHead{CLIP Raw} & \suppTableHead{CLIP Aux.} & \suppTableHead{All}\\
\midrule
\multirow{5}{*}{Public} & Fast-WAM & 0.660 & 0.359 & 0.290 & 0.271 & 0.353 & 0.436\\
 & Long-WAM & 0.695 & 0.405 & 0.354 & 0.271 & 0.354 & 0.485\\
 & LingBot-VA & 0.704 & 0.399 & 0.409 & 0.276 & 0.381 & 0.504\\
 & Cosmos3 & 0.715 & 0.371 & 0.304 & 0.291 & 0.454 & 0.463\\
 & GR00T & 0.944 & 0.547 & 0.489 & 0.311 & 0.553 & 0.660\\
\midrule
\multirow{5}{*}{Constructed} & Fast-WAM & 0.878 & 0.265 & 0.438 & 0.276 & 0.382 & 0.526\\
 & Long-WAM & 0.858 & 0.328 & 0.448 & 0.278 & 0.388 & 0.542\\
 & LingBot-VA & 0.868 & 0.345 & 0.531 & 0.283 & 0.416 & 0.580\\
 & Cosmos3 & 0.564 & 0.228 & 0.228 & 0.258 & 0.291 & 0.340\\
 & GR00T & 0.827 & 0.549 & 0.344 & 0.281 & 0.405 & 0.574\\
\bottomrule
\end{tabular*}
\end{table}

\FloatBarrier
% Unified subset- and family-level sub-metric breakdowns integrated into Appendix E.

\subsection{Subset-Level Physical, Geometric, and Interaction Criteria}
\label{sec:subset_detailed_results}
Tables~\ref{tab:subset_public_physics}, \ref{tab:subset_physical_rules}, \ref{tab:subset_public_geometry}, and~\ref{tab:subset_public_interaction} decompose all physical rules, geometric criteria, and interaction criteria separately across each model's 400 public episodes and 100 constructed episodes, reporting per-case aggregated means. Across all 2,500 rollouts, the six thermodynamic rules and three non-hardness material rules (color mixing, solubility, and combustibility) have zero triggered episodes, so the active material-property group score equals the hardness rule score on every applicable case.

\paragraph{Physical-rule shifts across subsets.}
Tables~\ref{tab:subset_public_physics} and~\ref{tab:subset_physical_rules} explain the opposing subset trends in $P$ observed in Table~\ref{tab:curated_results}. On public episodes, dual-arm models are penalized primarily by multi-object collisions under Impact ($0.329$--$0.493$ across $162$--$163$ cases) and gripper--object interpenetration under Hardness ($0.599$--$0.707$ across $241$--$242$ cases), whereas on constructed household tasks their multi-view inputs maintain high Impact ($0.881$--$0.942$) and Hardness ($0.855$--$0.950$) compliance. Conversely, state-free single-view Cosmos3 drops from $0.715$ to $0.564$ on constructed physics mainly because open-loop video generation warps rigid objects upon contact (Hardness falling from $0.622$ to $0.367$ across $75$ cases). Even simulator-executed GR00T drops on Impact ($0.835$ on $41$ public cases to $0.588$ on $46$ constructed cases) when open-loop 8-step action chunks strike articulated doors, hooks, or tools abruptly.

\paragraph{Geometric and interaction criteria across subsets.}
Tables~\ref{tab:subset_public_geometry} and~\ref{tab:subset_public_interaction} reveal why visual frame quality and continuous trajectory fidelity diverge in Section~\ref{sec:main_results}. Moving from public to constructed cases, GR00T's local visual quality increases from $0.833$ to $0.931$ under live SAPIEN rendering while $G_{\mathrm{base}}$ remains unchanged at $0.650$; however, its transition score $I_{\mathrm{trans}}$ plunges from $0.276$ ($152$ multi-chunk cases) to $0.105$ ($100$ multi-chunk cases) and local instruction alignment drops from $0.474$ to $0.337$, pulling $I$ down from $0.489$ to $0.344$. All $100$ constructed cases per model span multiple chunks ($K>1$), whereas $99/400$ dual-arm and $247$--$248/400$ single-arm public episodes have $K=1$ ($301$ and $153/152$ multi-chunk public cases, matching Table~\ref{tab:applicable_counts}), so single-chunk weight renormalization (Appendix~\ref{sec:supp_track_integration}) primarily affects single-arm public means. Meanwhile, LingBot-VA leads dual-arm transition ($0.496$ public, $0.711$ constructed) and global interaction ($0.506$ public, $0.706$ constructed) across both regimes, whereas Cosmos3 suffers severe camera-trajectory drift on longer constructed rollouts ($0.305$ to $0.107$).

\begin{table}[!ht]
\centering\small
\renewcommand{\arraystretch}{1.08}
\setlength{\tabcolsep}{5pt}
\caption{\textbf{Physical dimension scores by evaluation subset.} An en dash indicates no applicable episodes.}
\label{tab:subset_public_physics}
\label{tab:subset_constructed_physics}
\begin{tabular*}{\textwidth}{@{\extracolsep{\fill}}llcccc@{}}
\toprule
\suppTableHead{Subset} & \suppTableHead{Model} & \suppTableHead{Overall} & \suppTableHead{Mechanics} & \suppTableHead{Thermodynamics} & \suppTableHead{Material}\\
\midrule
\multirow{5}{*}{Public} & Fast-WAM & 0.660 & 0.672 & \textemdash & 0.599\\
 & Long-WAM & 0.695 & 0.723 & \textemdash & 0.644\\
 & LingBot-VA & 0.704 & 0.713 & \textemdash & 0.707\\
 & Cosmos3 & 0.715 & 0.783 & \textemdash & 0.622\\
 & GR00T & 0.944 & 0.939 & \textemdash & 0.948\\
\midrule
\multirow{5}{*}{Constructed} & Fast-WAM & 0.878 & 0.871 & \textemdash & 0.950\\
 & Long-WAM & 0.858 & 0.861 & \textemdash & 0.913\\
 & LingBot-VA & 0.868 & 0.883 & \textemdash & 0.855\\
 & Cosmos3 & 0.564 & 0.737 & \textemdash & 0.367\\
 & GR00T & 0.827 & 0.809 & \textemdash & 0.862\\
\bottomrule
\end{tabular*}
\end{table}

\begin{table}[!ht]
\edef\physicalRuleTableNumber{\thetable a}
\edef\physicalRuleTableAnchor{\theHtable.a}
\renewcommand{\thetable}{\physicalRuleTableNumber}
\renewcommand{\theHtable}{\physicalRuleTableAnchor}
\centering\footnotesize
\renewcommand{\arraystretch}{1.08}
\setlength{\tabcolsep}{2pt}
\caption{\textbf{Physical question scores by evaluation subset.} An en dash indicates no applicable episodes. Table~\ref{tab:applicable_counts} gives applicable counts.}
\label{tab:subset_physical_rules}
\begin{tabular*}{\textwidth}{@{\extracolsep{\fill}}ll*{14}{c}@{}}
\toprule
\suppTableHead{Subset} & \suppTableHead{Model} & \suppTableHead{Grav.} & \suppTableHead{Buoy.} & \suppTableHead{Comp.} & \suppTableHead{Impact} & \suppTableHead{Melt} & \suppTableHead{Sub.} & \suppTableHead{Vap.} & \suppTableHead{Cond.} & \suppTableHead{Dep.} & \suppTableHead{Freez.} & \suppTableHead{Color} & \suppTableHead{Sol.} & \suppTableHead{Hard.} & \suppTableHead{Comb.}\\
\midrule
\multirow{5}{*}{Public} & Fast-WAM & 0.674 & 0.980 & 0.765 & 0.329 & \textemdash & \textemdash & \textemdash & \textemdash & \textemdash & \textemdash & \textemdash & \textemdash & 0.599 & \textemdash\\
 & Long-WAM & 0.727 & 0.950 & 0.753 & 0.493 & \textemdash & \textemdash & \textemdash & \textemdash & \textemdash & \textemdash & \textemdash & \textemdash & 0.644 & \textemdash\\
 & LingBot-VA & 0.717 & 0.980 & 0.915 & 0.441 & \textemdash & \textemdash & \textemdash & \textemdash & \textemdash & \textemdash & \textemdash & \textemdash & 0.707 & \textemdash\\
 & Cosmos3 & 0.784 & \textemdash & \textemdash & 0.525 & \textemdash & \textemdash & \textemdash & \textemdash & \textemdash & \textemdash & \textemdash & \textemdash & 0.622 & \textemdash\\
 & GR00T & 0.939 & \textemdash & \textemdash & 0.835 & \textemdash & \textemdash & \textemdash & \textemdash & \textemdash & \textemdash & \textemdash & \textemdash & 0.948 & \textemdash\\
\midrule
\multirow{5}{*}{Constructed} & Fast-WAM & 0.875 & \textemdash & 0.900 & 0.942 & \textemdash & \textemdash & \textemdash & \textemdash & \textemdash & \textemdash & \textemdash & \textemdash & 0.950 & \textemdash\\
 & Long-WAM & 0.865 & \textemdash & 0.900 & 0.882 & \textemdash & \textemdash & \textemdash & \textemdash & \textemdash & \textemdash & \textemdash & \textemdash & 0.913 & \textemdash\\
 & LingBot-VA & 0.887 & \textemdash & 0.855 & 0.881 & \textemdash & \textemdash & \textemdash & \textemdash & \textemdash & \textemdash & \textemdash & \textemdash & 0.855 & \textemdash\\
 & Cosmos3 & 0.758 & \textemdash & 0.550 & 0.577 & \textemdash & \textemdash & \textemdash & \textemdash & \textemdash & \textemdash & \textemdash & \textemdash & 0.367 & \textemdash\\
 & GR00T & 0.878 & \textemdash & 0.912 & 0.588 & \textemdash & \textemdash & \textemdash & \textemdash & \textemdash & \textemdash & \textemdash & \textemdash & 0.862 & \textemdash\\
\bottomrule
\end{tabular*}
\end{table}
\addtocounter{table}{-1}

\begin{table}[!ht]
\centering\small
\renewcommand{\arraystretch}{1.08}
\setlength{\tabcolsep}{4.5pt}
\caption{\textbf{Geometric criteria on public and constructed subsets.} 3D Cons.\ 1.0 is the inherited score. 3D Cons.\ includes Path.}
\label{tab:subset_public_geometry}
\label{tab:subset_constructed_geometry}
\begin{tabular*}{\textwidth}{@{\extracolsep{\fill}}llrrrrrr@{}}
\toprule
\suppTableHead{Subset} & \suppTableHead{Model} & \suppTableHead{GS} & \suppTableHead{Meta} & \suppTableHead{Camera Motion} & \suppTableHead{3D Cons. 1.0} & \suppTableHead{Path} & \suppTableHead{3D Cons.}\\
\midrule
\multirow{5}{*}{Public} & Fast-WAM & 0.328 & 0.338 & 0.511 & 0.392 & 0.260 & 0.359\\
 & Long-WAM & 0.382 & 0.381 & 0.609 & 0.457 & 0.250 & 0.405\\
 & LingBot-VA & 0.384 & 0.388 & 0.563 & 0.445 & 0.261 & 0.399\\
 & Cosmos3 & 0.388 & 0.357 & 0.305 & 0.350 & 0.434 & 0.371\\
 & GR00T & 0.532 & 0.498 & 0.919 & 0.650 & 0.237 & 0.547\\
\midrule
\multirow{5}{*}{Constructed} & Fast-WAM & 0.264 & 0.249 & 0.378 & 0.297 & 0.171 & 0.265\\
 & Long-WAM & 0.300 & 0.292 & 0.507 & 0.366 & 0.214 & 0.328\\
 & LingBot-VA & 0.334 & 0.335 & 0.526 & 0.398 & 0.184 & 0.345\\
 & Cosmos3 & 0.310 & 0.321 & 0.107 & 0.246 & 0.174 & 0.228\\
 & GR00T & 0.557 & 0.557 & 0.836 & 0.650 & 0.246 & 0.549\\
\bottomrule
\end{tabular*}
\end{table}

\begin{table}[!ht]
\centering\scriptsize
\renewcommand{\arraystretch}{1.10}
\setlength{\tabcolsep}{1.5pt}
\caption{\textbf{Interaction criteria on public and constructed subsets.} CLIP Raw is cosine similarity from $-1$ to $1$. CLIP Aux.\ is calibrated. Interact.\ 1.0 is the inherited score. Interact.\ includes Vel. See Appendix~\ref{sec:supp_track_integration} for single-chunk reweighting.}
\label{tab:synthetic_submetrics}
\label{tab:subset_public_interaction}
\label{tab:subset_constructed_interaction}
\begin{tabular*}{\textwidth}{@{\extracolsep{\fill}}llrrrrrrrrrrrr@{}}
\toprule
\suppTableHead{Subset} & \suppTableHead{Model} & \suppTableHead{Visual} & \suppTableHead{Instr.} & \suppTableHead{Chunk} & \suppTableHead{Trans.} & \suppTableHead{Global} & \suppTableHead{Long Range} & \suppTableHead{Global Text} & \suppTableHead{CLIP Raw} & \suppTableHead{CLIP Aux.} & \suppTableHead{Interact. 1.0} & \suppTableHead{Vel.} & \suppTableHead{Interact.}\\
\midrule
\multirow{5}{*}{Public} & Fast-WAM & 0.375 & 0.338 & 0.359 & 0.144 & 0.393 & 0.404 & 0.383 & 0.271 & 0.353 & 0.336 & 0.143 & 0.290\\
 & Long-WAM & 0.430 & 0.381 & 0.406 & 0.318 & 0.479 & 0.499 & 0.460 & 0.271 & 0.354 & 0.419 & 0.138 & 0.354\\
 & LingBot-VA & 0.505 & 0.411 & 0.459 & 0.496 & 0.506 & 0.527 & 0.484 & 0.276 & 0.381 & 0.485 & 0.144 & 0.409\\
 & Cosmos3 & 0.446 & 0.366 & 0.407 & 0.258 & 0.257 & 0.270 & 0.252 & 0.291 & 0.454 & 0.346 & 0.173 & 0.304\\
 & GR00T & 0.833 & 0.474 & 0.564 & 0.276 & 0.768 & 0.793 & 0.767 & 0.311 & 0.553 & 0.606 & 0.127 & 0.489\\
\midrule
\multirow{5}{*}{Constructed} & Fast-WAM & 0.423 & 0.362 & 0.392 & 0.546 & 0.584 & 0.616 & 0.552 & 0.276 & 0.382 & 0.495 & 0.211 & 0.438\\
 & Long-WAM & 0.427 & 0.355 & 0.391 & 0.523 & 0.630 & 0.656 & 0.604 & 0.278 & 0.388 & 0.502 & 0.231 & 0.448\\
 & LingBot-VA & 0.543 & 0.419 & 0.477 & 0.711 & 0.706 & 0.756 & 0.656 & 0.283 & 0.416 & 0.610 & 0.218 & 0.531\\
 & Cosmos3 & 0.368 & 0.313 & 0.343 & 0.199 & 0.126 & 0.136 & 0.116 & 0.258 & 0.291 & 0.230 & 0.220 & 0.228\\
 & GR00T & 0.931 & 0.337 & 0.399 & 0.105 & 0.608 & 0.630 & 0.606 & 0.281 & 0.405 & 0.374 & 0.224 & 0.344\\
\bottomrule
\end{tabular*}
\end{table}

\FloatBarrier
\subsection{Results by Constructed Task Family}
\label{sec:synthetic_component}
Each embodiment cohort spans five constructed families of 20 episodes each (Table~\ref{tab:synthetic_families}): the three dual-arm models share 100 ARX5 episodes, while Cosmos3 and GR00T share 100 single-arm Panda episodes. Decomposing each family across all three tracks and both CAE metrics highlights four task-level mechanisms supporting Sections~\ref{sec:experimental_setup}--\ref{sec:harness_results}:
\begin{enumerate}
\setlength{\itemsep}{3pt}\setlength{\topsep}{3pt}
\item \textbf{Origin of the 14 N/A physical runs.} All 14 constructed runs without a triggered physical rule ($4$ Fast-WAM, $5$ Long-WAM, and $5$ LingBot-VA, leaving $16, 15, 15$ applicable cases) arise in dual-arm \emph{cabinet articulation}, where smooth horizontal door and drawer pulls involve neither free-fall nor object deformation.
\item \textbf{Post-release precision in cup hanging.} \emph{Cup hanging} is the hardest family for dual-arm models (All: $0.372$--$0.537$) due to tight handle-to-hook clearance: slipped mugs lower both $P$ ($0.617$--$0.760$) and $I$ ($0.290$--$0.482$). By contrast, GR00T retains high $P$ ($0.807$) and $G$ ($0.596$) yet drops to $I=0.137$ because missed mugs fall realistically under PyBullet gravity while failing the hanging goal, whereas Cosmos3 scores $P=0.926$ with $\text{Path}=0.100$ by leaving the mug nearly stationary.
\item \textbf{Non-contact tool swings exposed by CAE.} In \emph{tool / dowel manipulation}, dual-arm models obtain high $P$ ($0.871$--$0.941$) and moderate MLLM interaction ratings ($I=0.420$--$0.461$) by swinging the tool toward the target, yet object-centric CAE path fidelity drops to $0.007$--$0.076$ when missed contact leaves the dowel unmoved; closed-loop GR00T physically displaces the dowel and achieves its highest path ($0.351$) and velocity ($0.270$) scores.
\item \textbf{History conditioning in multi-stage reorientation.} On \emph{cup uprighting} and \emph{cup hanging}, Long-WAM leads all dual-arm models in path fidelity ($0.290$ and $0.327$), reflecting its 48-step observation history across pick, reorient, and placement phases.
\end{enumerate}

\begin{table}[!ht]
\centering\small
\renewcommand{\arraystretch}{1.04}
\setlength{\tabcolsep}{4pt}
\caption{\textbf{Track and CAE scores by constructed task family.} Each family contains 20 episodes per model within the ARX5 dual-arm and Panda single-arm cohorts. All averages the three tracks over cases with a triggered physical rule.}
\label{tab:synthetic_families}
\label{tab:subset_constructed_family_tracks}
\begin{tabular*}{\textwidth}{@{\extracolsep{\fill}}llrrrrrr@{}}
\toprule
\suppTableHead{Task family} & \suppTableHead{Model} & \suppTableHead{Physical} & \suppTableHead{3D Cons.} & \suppTableHead{Interact.} & \suppTableHead{All} & \suppTableHead{Path} & \suppTableHead{Vel.}\\
\midrule
\multirow{5}{*}{Cabinet articulation}
 & Fast-WAM & 0.950 & 0.311 & 0.523 & 0.598 & 0.265 & 0.269\\
 & Long-WAM & 0.950 & 0.345 & 0.558 & 0.616 & 0.233 & 0.215\\
 & LingBot-VA & 0.950 & 0.317 & 0.620 & 0.628 & 0.263 & 0.188\\
 & Cosmos3 & 0.594 & 0.212 & 0.179 & 0.328 & 0.298 & 0.228\\
 & GR00T & 0.879 & 0.519 & 0.357 & 0.585 & 0.116 & 0.213\\
\midrule
\multirow{5}{*}{Cup uprighting}
 & Fast-WAM & 0.948 & 0.268 & 0.384 & 0.533 & 0.200 & 0.198\\
 & Long-WAM & 0.855 & 0.317 & 0.398 & 0.523 & 0.290 & 0.274\\
 & LingBot-VA & 0.902 & 0.399 & 0.490 & 0.597 & 0.257 & 0.206\\
 & Cosmos3 & 0.329 & 0.198 & 0.234 & 0.254 & 0.086 & 0.240\\
 & GR00T & 0.803 & 0.584 & 0.471 & 0.620 & 0.288 & 0.201\\
\midrule
\multirow{5}{*}{Cup hanging}
 & Fast-WAM & 0.617 & 0.183 & 0.315 & 0.372 & 0.156 & 0.202\\
 & Long-WAM & 0.665 & 0.278 & 0.290 & 0.411 & 0.327 & 0.243\\
 & LingBot-VA & 0.760 & 0.370 & 0.482 & 0.537 & 0.254 & 0.208\\
 & Cosmos3 & 0.926 & 0.210 & 0.248 & 0.462 & 0.100 & 0.273\\
 & GR00T & 0.807 & 0.596 & 0.137 & 0.513 & 0.186 & 0.186\\
\midrule
\multirow{5}{*}{Tool / dowel manipulation}
 & Fast-WAM & 0.941 & 0.233 & 0.439 & 0.537 & 0.076 & 0.136\\
 & Long-WAM & 0.894 & 0.318 & 0.420 & 0.544 & 0.059 & 0.169\\
 & LingBot-VA & 0.871 & 0.256 & 0.461 & 0.529 & 0.007 & 0.240\\
 & Cosmos3 & 0.417 & 0.309 & 0.198 & 0.308 & 0.281 & 0.229\\
 & GR00T & 0.792 & 0.484 & 0.500 & 0.592 & 0.351 & 0.270\\
\midrule
\multirow{5}{*}{Tissue-box pushing}
 & Fast-WAM & 0.950 & 0.333 & 0.528 & 0.604 & 0.156 & 0.248\\
 & Long-WAM & 0.950 & 0.383 & 0.572 & 0.635 & 0.163 & 0.255\\
 & LingBot-VA & 0.878 & 0.381 & 0.603 & 0.621 & 0.138 & 0.249\\
 & Cosmos3 & 0.556 & 0.211 & 0.278 & 0.348 & 0.104 & 0.127\\
 & GR00T & 0.855 & 0.563 & 0.255 & 0.558 & 0.291 & 0.250\\
\bottomrule
\end{tabular*}
\end{table}

\clearpage
\section{Qualitative Demonstrations}
\label{sec:supp_demos}
Table~\ref{tab:supp_demo_index} indexes the constructed simulation demonstrations, public stacking episodes, and newly collected AgileX Piper bimanual sequences in Appendix~\ref{sec:supp_demos}.

\begin{table}[!ht]
\centering\small
\setlength{\tabcolsep}{4pt}
\renewcommand{\arraystretch}{1.10}
\caption{\textbf{Index of qualitative demonstrations.} Overview of constructed references, cross-model rollout comparisons, and real-robot bimanual episodes in Appendix~\ref{sec:supp_demos}.}
\label{tab:supp_demo_index}
\begin{tabular*}{\textwidth}{@{\extracolsep{\fill}}>{\raggedright\arraybackslash}p{.23\textwidth}>{\raggedright\arraybackslash}p{.59\textwidth}c@{}}
\toprule
\suppTableHead{Category} & \suppTableHead{Tasks shown} & \suppTableHead{Figures}\\
\midrule
Constructed references & Jar pushing, mug hanging, hammer striking & \ref{fig:reference_gallery}\\
Public stacking episodes & Dual-arm block stacking, single-arm cube stacking & \ref{fig:rollouts1}, \ref{fig:rollouts3}\\
Articulated objects & Cabinet-door opening, drawer opening & \ref{fig:constructed_comparison_1}, \ref{fig:constructed_comparison_2}\\
Orientation and hanging & Cup uprighting, cup hanging & \ref{fig:constructed_comparison_3}, \ref{fig:constructed_comparison_4}\\
Tool use and pushing & Tool--dowel interaction, tissue-box pushing & \ref{fig:constructed_comparison_5}, \ref{fig:constructed_comparison_6}\\
Object placement & Apple placement, kettle placement & \ref{fig:legacy_apple_rollouts}, \ref{fig:legacy_kettle_rollouts}\\
Real-robot bimanual tasks & Utensil setup, tower building, wiping, packing, pouring, counting, boxing, stacking & \ref{fig:real_robot_tasks_first}, \ref{fig:real_robot_tasks_second}\\
\bottomrule
\end{tabular*}
\end{table}

\subsection{Constructed Canonical References}
\label{sec:supp_reference_demos}
\begin{figure}[H]
\centering
\includegraphics[width=.94\textwidth]{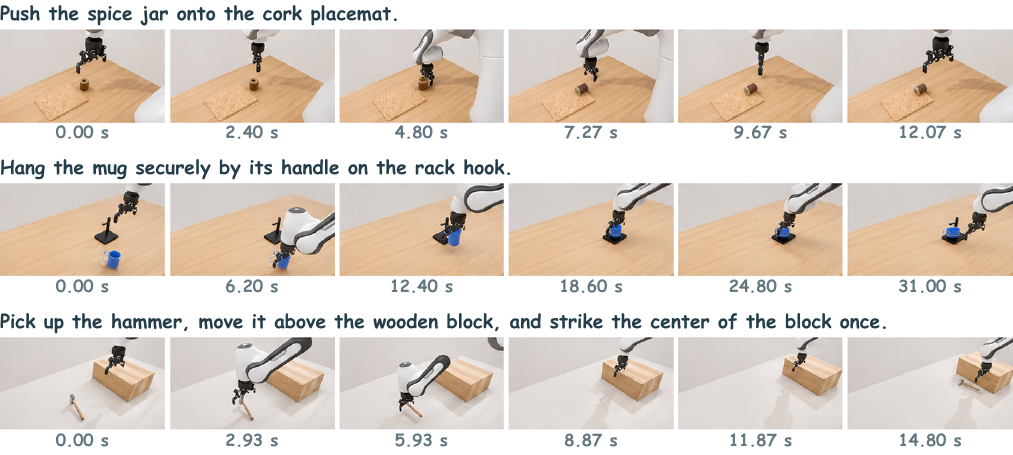}
\caption{\textbf{Constructed canonical references.} Physics-checked simulation references for household pushing, hanging, and tool use across six timestamps.}
\label{fig:demos}
\label{fig:reference_gallery}
\end{figure}

Figure~\ref{fig:reference_gallery} shows three physics-checked constructed canonical references for household pushing, hanging, and tool striking across six timestamps.

\clearpage
\subsection{Public Episodes: Block and Cube Stacking}
\label{sec:supp_stacking_demos}
Dual-arm block stacking (RoboDojo) and single-arm cube stacking (ManiSkill3) compare multi-object layout preservation and post-release support across native rollout protocols.

\begin{figure}[H]
\centering
\includegraphics[width=.96\textwidth]{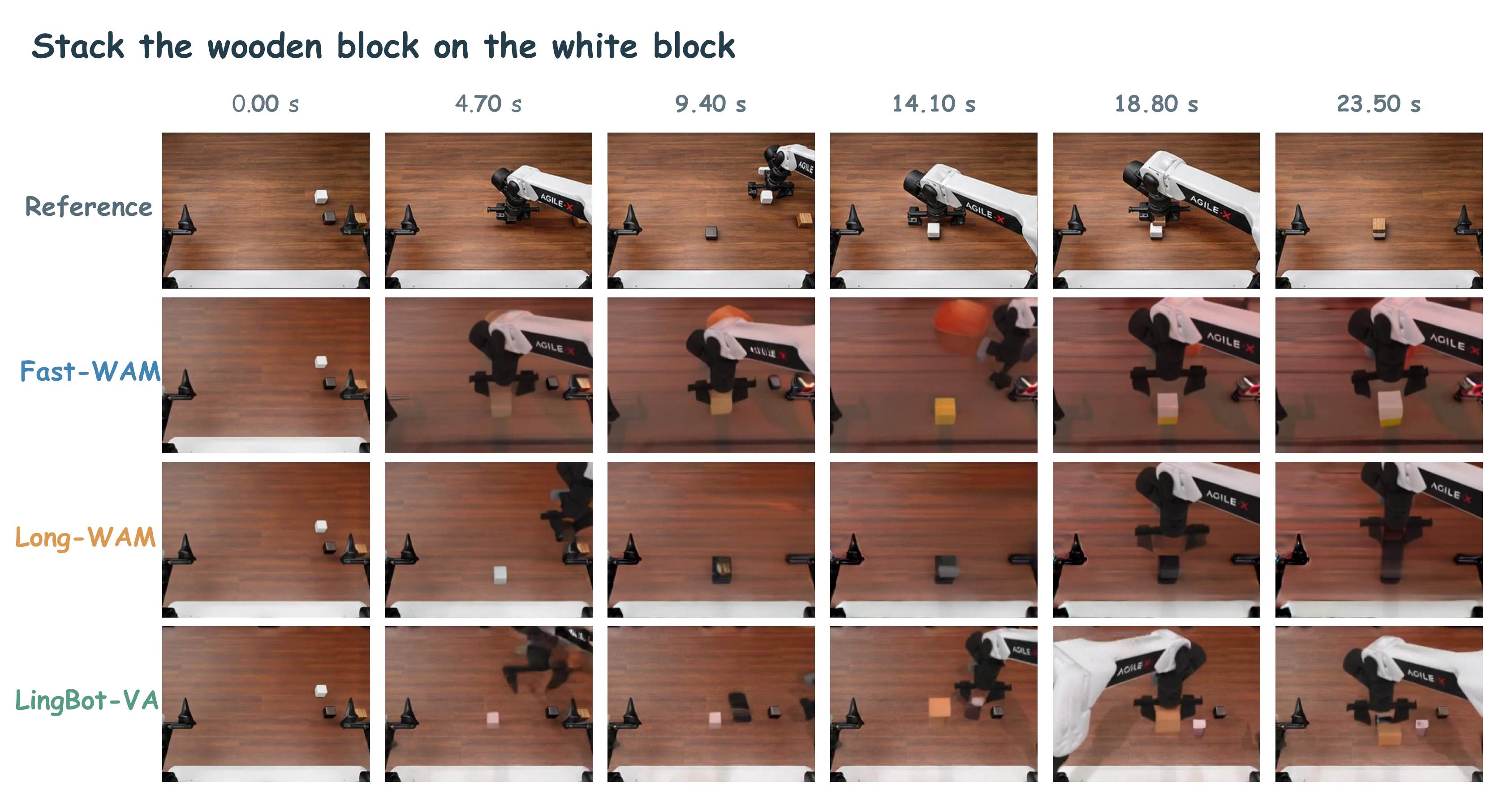}
\caption{\textbf{Block stacking (dual-arm public episode).} Reference and Fast-WAM, Long-WAM, and LingBot-VA rollouts on a shared RoboDojo episode.}
\label{fig:rollouts1}
\end{figure}
\vspace{4pt}
\begin{figure}[H]
\centering
\includegraphics[width=.96\textwidth]{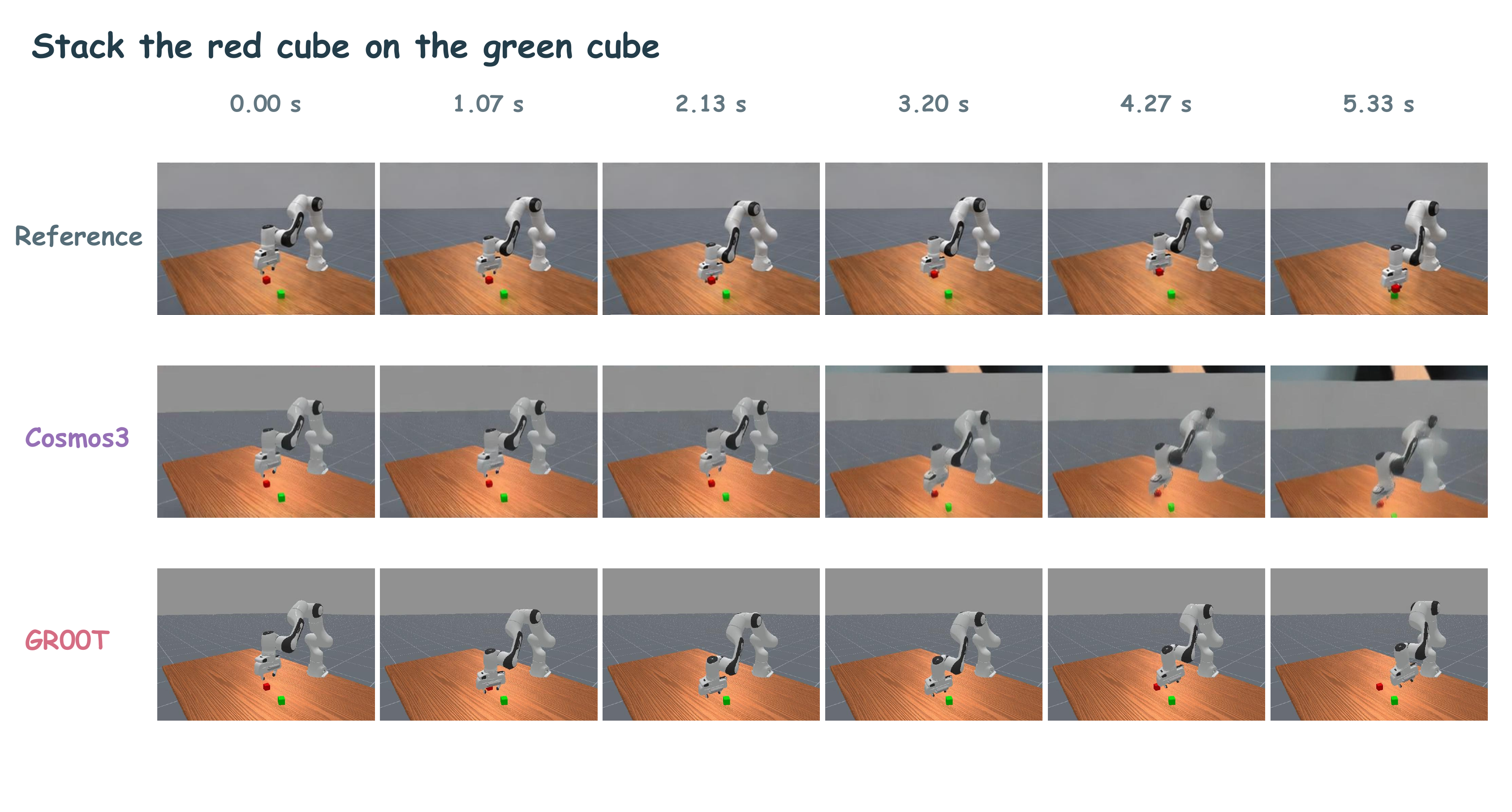}
\caption{\textbf{Cube stacking (single-arm public episode).} Reference, Cosmos3, and GR00T rollouts at six matched timestamps on a shared ManiSkill3 episode.}
\label{fig:rollouts3}
\end{figure}

\clearpage
\subsection{Constructed Tasks: Cabinet Door and Drawer Opening}
\label{sec:supp_articulation_demos}
Cabinet-door and drawer manipulation test revolute and prismatic joint articulation under physical constraints. Cosmos3 generates video rollouts, and GR00T executes actions in simulation.

\begin{figure}[H]
\centering
\includegraphics[width=\textwidth]{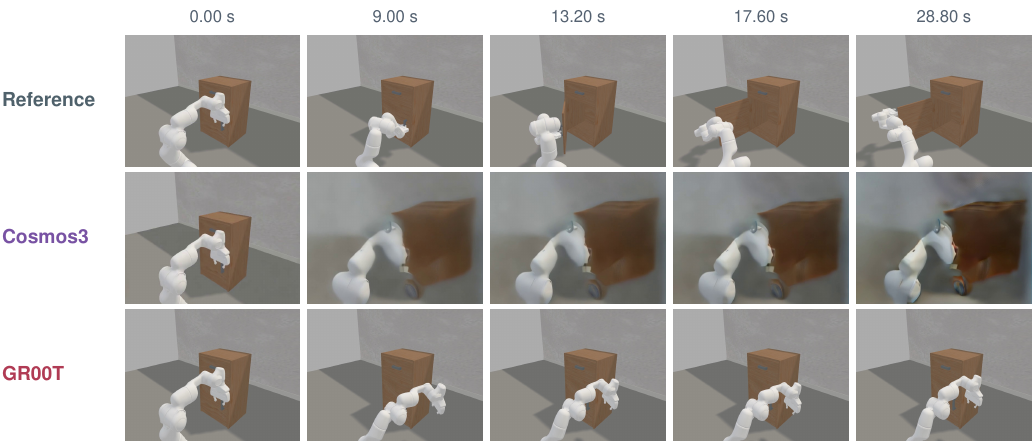}
\caption{\textbf{Cabinet-door opening.} Panda reference, Cosmos3, and GR00T at five matched elapsed times.}
\label{fig:constructed_comparison_1}
\end{figure}
\vspace{12pt}
\begin{figure}[H]
\centering
\includegraphics[width=\textwidth]{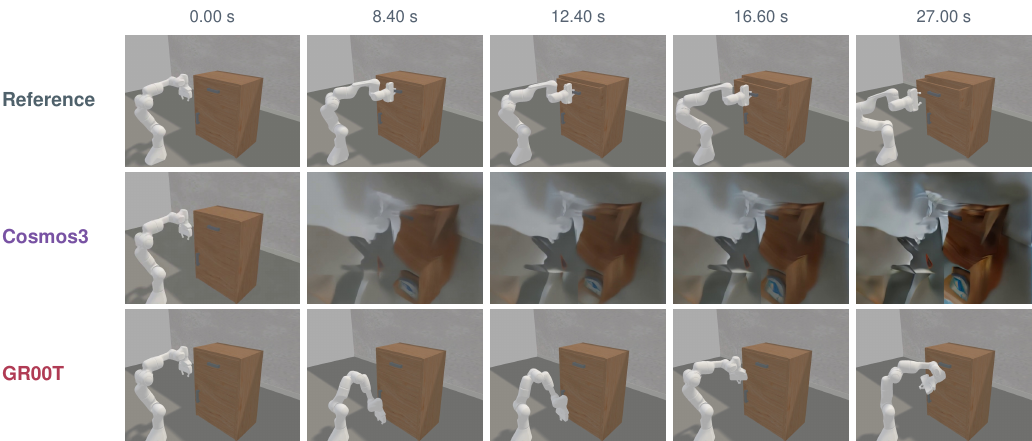}
\caption{\textbf{Drawer opening.} Panda reference, Cosmos3, and GR00T at five matched elapsed times.}
\label{fig:constructed_comparison_2}
\end{figure}

\clearpage
\subsection{Constructed Tasks: Cup Uprighting and Hanging}
\label{sec:supp_pose_demos}
Cup uprighting evaluates 3D orientation control, whereas cup hanging additionally requires precise handle-to-hook clearance and post-release support.

\begin{figure}[H]
\centering
\includegraphics[width=\textwidth]{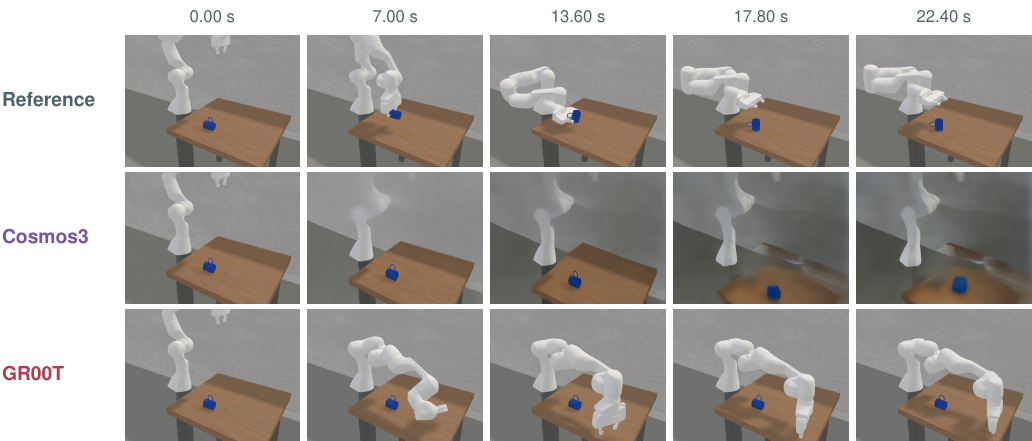}
\caption{\textbf{Cup uprighting.} Panda reference, Cosmos3, and GR00T at five matched elapsed times.}
\label{fig:constructed_comparison_3}
\end{figure}
\vspace{12pt}
\begin{figure}[H]
\centering
\includegraphics[width=\textwidth]{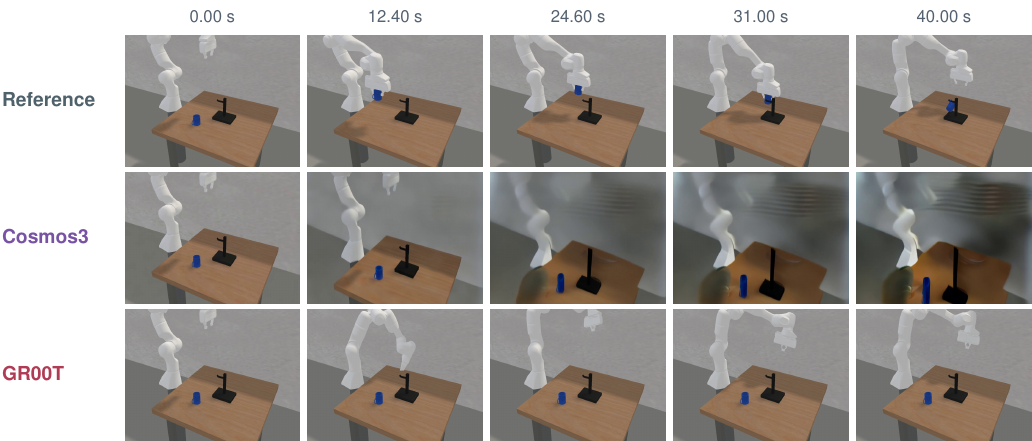}
\caption{\textbf{Cup hanging.} Panda reference, Cosmos3, and GR00T at five matched elapsed times.}
\label{fig:constructed_comparison_4}
\end{figure}

\clearpage
\subsection{Constructed Tasks: Tool Use and Tissue-Box Pushing}
\label{sec:supp_contact_demos}
Tool--dowel striking and tissue-box pushing evaluate non-prehensile, contact-driven object motion.

\begin{figure}[H]
\centering
\includegraphics[width=\textwidth]{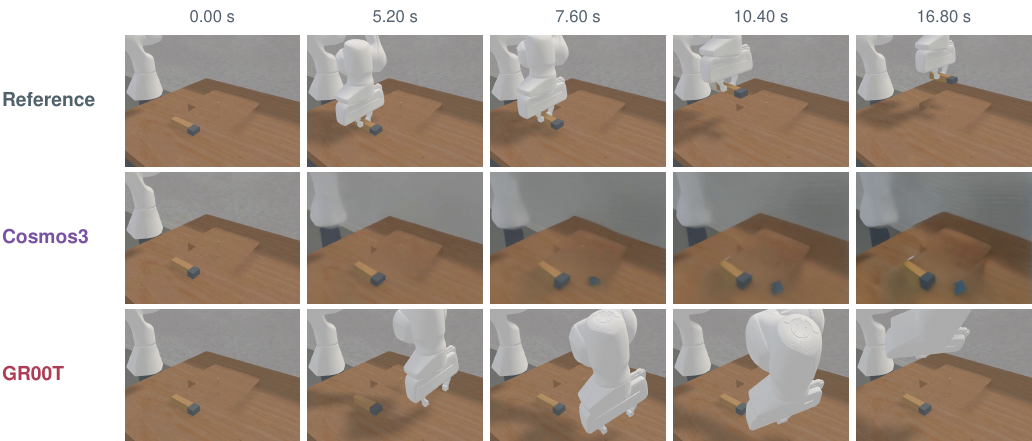}
\caption{\textbf{Tool--dowel interaction.} Panda reference, Cosmos3, and GR00T at five matched elapsed times.}
\label{fig:constructed_comparison_5}
\end{figure}
\vspace{12pt}
\begin{figure}[H]
\centering
\includegraphics[width=\textwidth]{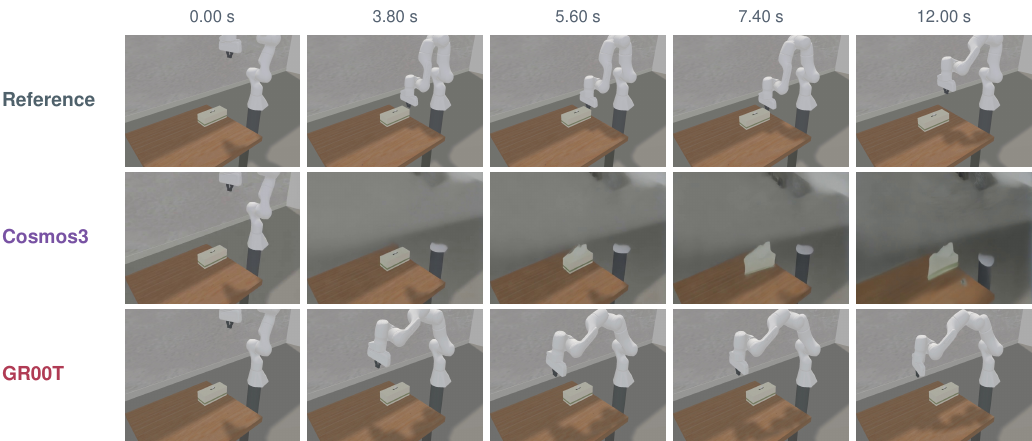}
\caption{\textbf{Tissue-box pushing.} Panda reference, Cosmos3, and GR00T at five matched elapsed times.}
\label{fig:constructed_comparison_6}
\end{figure}

\clearpage
\subsection{Constructed Tasks: Apple and Kettle Placement}
\label{sec:supp_original_comparisons}
Apple and kettle placement illustrate how canonical tasks adapt across dual-arm head-view and single-arm DROID camera interfaces. Table~\ref{tab:synthetic_families} summarizes the five task families used in quantitative evaluation.

\begin{figure}[H]
\centering
\includegraphics[width=\textwidth]{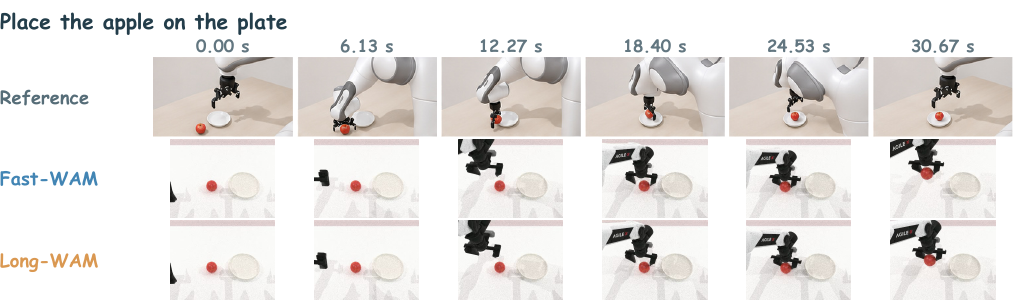}
\caption{\textbf{Apple placement.} Constructed reference with Fast-WAM and Long-WAM rollouts across six timestamps.}
\label{fig:legacy_apple_rollouts}
\end{figure}
\vspace{12pt}
\begin{figure}[H]
\centering
\includegraphics[width=\textwidth]{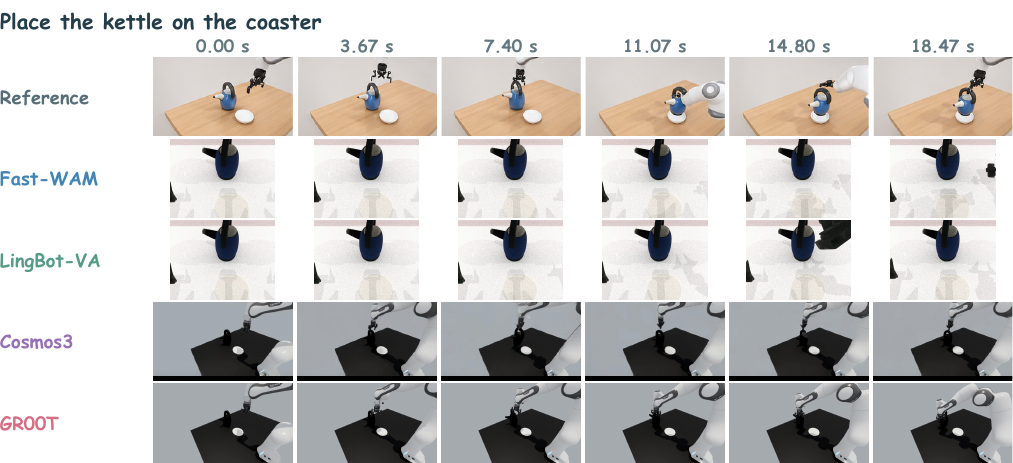}
\caption{\textbf{Kettle placement.} Constructed reference and four model rollouts across six timestamps.}
\label{fig:legacy_kettle_rollouts}
\end{figure}

\clearpage
\subsection{Real-Robot Bimanual Demonstrations}
\label{sec:real_robot_demonstrations}
\begin{figure}[H]
\centering
\includegraphics[width=.97\textwidth]{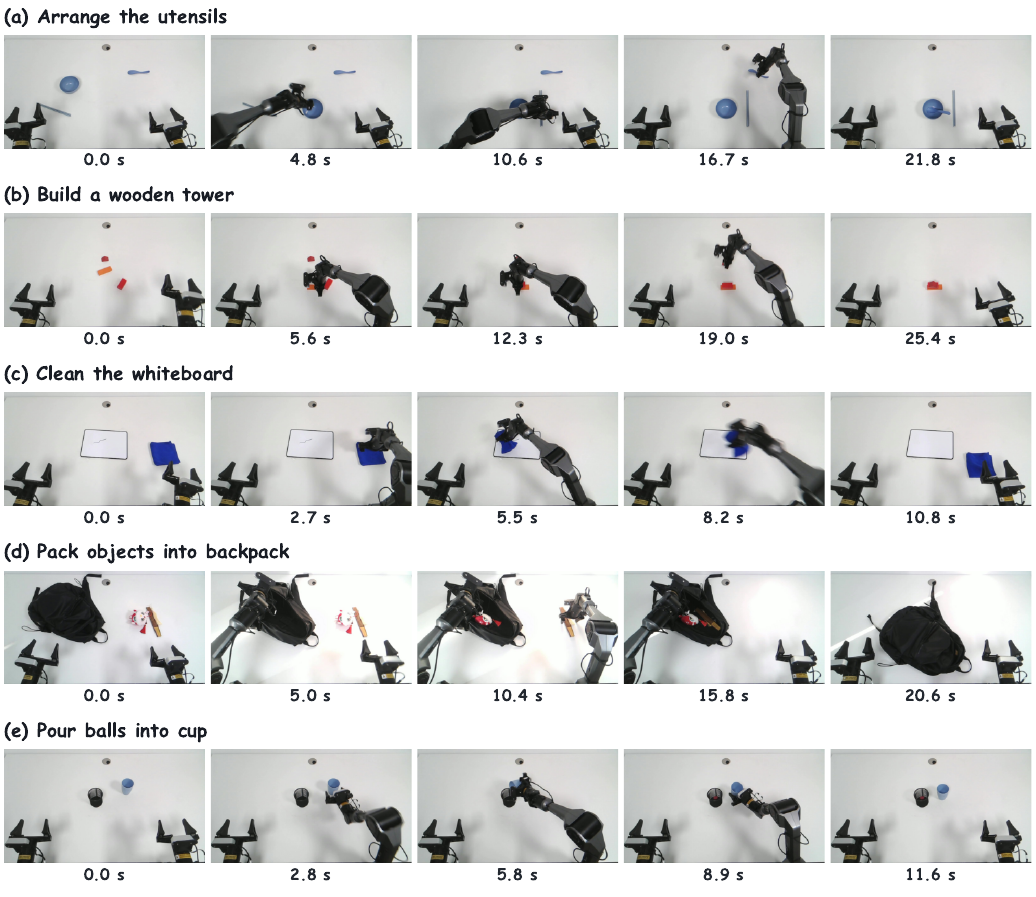}
\caption{\textbf{Real-robot bimanual demonstrations (a--e).} Five front-camera frames with elapsed timestamps and task instructions per episode: \textbf{(a)}~\emph{``Place the bowl near the front edge of the table, centered from left to right. Put the spoon inside the bowl and place the chopsticks to the right of the bowl''}; \textbf{(b)}~\emph{``Build a tower with the wooden blocks. Place the longest block at the bottom and the triangular roof block at the top''}; \textbf{(c)}~\emph{``Pick up the towel and wipe the writing off the whiteboard''}; \textbf{(d)}~\emph{``Put the objects on the table into the backpack''}; and \textbf{(e)}~\emph{``Pour the balls from the blue cup into the other cup.''}}
\label{fig:real_robot_tasks_first}
\end{figure}

\begin{figure}[H]
\centering
\includegraphics[width=.97\textwidth]{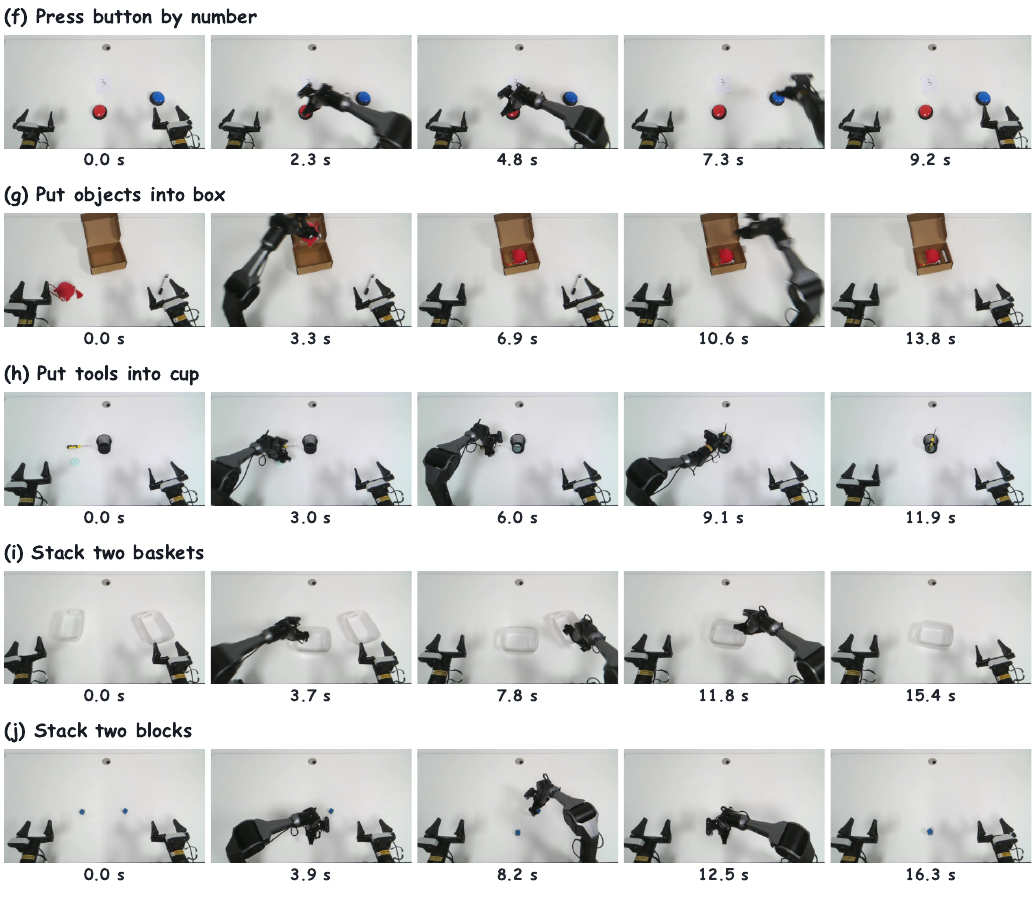}
\caption{\textbf{Real-robot bimanual demonstrations (f--j).} Five front-camera frames with elapsed timestamps and task instructions per episode: \textbf{(f)}~\emph{``Read the number on the piece of paper on the table. Press the red button that many times, then press the blue button once''}; \textbf{(g)}~\emph{``Put the objects on the table into the cardboard box''}; \textbf{(h)}~\emph{``Put the tools on the table, including the screwdriver and pens, into the cup''}; \textbf{(i)}~\emph{``Stack the baskets on top of one another''}; and \textbf{(j)}~\emph{``Stack the blocks vertically, one on top of another.''}}
\label{fig:real_robot_tasks_second}
\end{figure}

Figures~\ref{fig:real_robot_tasks_first} and~\ref{fig:real_robot_tasks_second} present ten newly collected real-robot bimanual demonstrations recorded on a dual-arm AgileX Piper setup (accompanying the supplementary rollout videos and held separate from the 1,100-case quantitative pool in Table~\ref{tab:data_sources}), spanning utensil arrangement, multi-block tower building, whiteboard wiping, deformable backpack packing, ball pouring, count-conditioned button pressing, object boxing, tool placement, and basket/block stacking across varied durations (9.3--25.7\,s).

\clearpage
\FloatBarrier
\end{document}